\documentclass{article} 
\usepackage{nips15submit_e,times}

\usepackage{graphicx}
\usepackage{amsmath,amssymb} 
\usepackage{color}

\usepackage[utf8]{inputenc}
\usepackage{times}
\usepackage{epsfig}

\usepackage{url}

\usepackage{enumerate}
\usepackage{stfloats}

\usepackage[toc,page]{appendix}
\usepackage[font=small,labelfont=bf]{caption}

\usepackage{xcolor}
\usepackage{amsfonts}
\usepackage[vlined, ruled]{algorithm2e}
\usepackage{caption}
\usepackage{subcaption}
\usepackage{bm}
\usepackage{isomath}
\usepackage{algorithm2e}
\usepackage[numbers]{natbib}
\usepackage{bbold}
\usepackage{dsfont}

\usepackage{footmisc}

\nipsfinaltrue

\usepackage{multirow}
\def\ie{\emph{i.e.}}
\def\etc{\emph{etc}}

\renewcommand\vec[1]{\ensuremath\boldsymbol{#1}}
\renewcommand\cdots{...}

\newcommand{\tY}{\vec{\mathcal{Y}}}

\newcommand{\vy}{\mathbf{y}}
\newcommand{\valpha}{\bm{\alpha}}
\newcommand{\tX}{\vec{\mathcal{X}}}

\newcommand{\mX}{\mathbf{X}}
\newcommand{\vx}{\mathbf{x}}

\newcommand{\mbr}[1]{\mathbb{R}^{#1}}

\newcommand{\vbeta}{\vec{\beta}}
\newcommand{\vv}{\mathbf{v}}
\newcommand{\tV}{\vec{\mathcal{V}}}
\newcommand{\tVnb}{\mathcal{V}}
\newcommand{\tXnb}{\mathcal{X}}
\newcommand{\tYnb}{\mathcal{Y}}
\newcommand{\tE}{\vec{\mathcal{E}}}
\newcommand{\tEH}{\vec{\mathcal{\hat{E}}}}
\newcommand{\tVH}{\vec{\mathcal{\bar{V}}}}
\newcommand{\tVT}{\vec{\mathcal{\hat{V}}}}
\newcommand{\idx}[1]{\mathcal{I}_{#1}}

\newcommand{\semipd}[1]{\mathcal{S}_{+}^{#1}}

\newcommand{\vu}{\mathbf{u}}

\newcommand{\vub}{\mathbf{\bar{u}}}
\newcommand{\vz}{\mathbf{z}}
\newcommand{\vzeta}{\boldsymbol{\zeta}}

\newcommand{\vphi}{\boldsymbol{\phi}}

\newcommand{\bigoh}{\mathcal{O}}

\newcommand{\vj}{\vec{j}}

\newcommand{\enorm}[1]{\left\|{#1}\right\|_2}
\newcommand{\fnorm}[1]{\left\|{#1}\right\|_F}

\newcommand{\set}[1]{\left\{#1\right\}}

\DeclareMathOperator*{\kronstack}{\uparrow\!\otimes}

\DeclareMathOperator*{\sgn}{Sgn}
\DeclareMathOperator*{\hosvd}{HOSVD}

\DeclareMathOperator*{\diag}{daig}

\newcommand{\suptensor}[1]{\mathfrak{S}^{d}}

\newcommand{\mLambda}{\bm{\lambda}}
\newcommand{\mU}{\bm{U}}
\newcommand{\mV}{\bm{V}}

\newcommand{\piA}{{\Pi_A}}
\newcommand{\piB}{{\Pi_B}}

\def\eg{\emph{e.g.}}

\newcommand{\mygthree}[1]{\boldsymbol{\mathcal{G}}\!\left(\!#1\!\right)}

\newcommand{\tG}{\boldsymbol{\mathcal{G}}}

\newcommand{\vi}{\mathbf{i}}

\newcommand{\comment}[1]{}

\newcommand{\idxJ}{\mathcal{J}}

\newcommand{\RaiseBiggBar}[2]{\ensuremath{\raisebox{#1}{$\Bigg|#2$}}}

\title{\Large Tensor Representations via Kernel Linearization for Action Recognition from 3D Skeletons (Extended Version)}

\author{Piotr Koniusz\textsuperscript{1,3}\qquad\qquad Anoop Cherian\textsuperscript{2,3}\qquad\qquad Fatih Porikli\textsuperscript{1,2,3}\\
$^1$NICTA/Data61/CSIRO, \\$^2$Australian Centre for Robotic Vision, $^3$Australian National University\\
firstname.lastname@\{data61.csiro.au, anu.edu.au\}
}

\newcommand\keywords[1]{}
\nipsfinalcopy 

\begin{document}

\maketitle

\begin{abstract}
In this paper, we explore tensor representations that can compactly capture higher-order relationships between skeleton joints for 3D action recognition. We first define RBF kernels on 3D joint sequences, which are then linearized to form kernel descriptors. The higher-order outer-products of these kernel descriptors form our tensor representations. We present two different kernels for action recognition, namely (i) a \emph{sequence compatibility kernel} that captures the spatio-temporal compatibility of joints in one sequence against those in the other, and (ii) a \emph{dynamics compatibility kernel} that explicitly models the action dynamics of a sequence. Tensors formed from these kernels are then used to train an SVM. We present experiments on several benchmark datasets and demonstrate state of the art results, substantiating the effectiveness of our representations.
\end{abstract}

\section{Introduction}
\label{sec:intro}
Human action recognition is a central problem in computer vision with potential impact in surveillance, human-robot interaction, elderly assistance systems, and gaming,  to name a few. While there have been significant advancements in this area over the past few years, action recognition in unconstrained settings still remains a challenge. There have been research to simplify the problem from using RGB cameras to more sophisticated sensors such as Microsoft Kinect that can localize human body-parts and produce moving 3D skeletons~\cite{shotton2013real}; these skeletons are then used for recognition. Unfortunately, these skeletons are often noisy due to the difficulty in localizing body-parts, self-occlusions, and  sensor range errors; thus necessitating higher-order reasoning on these 3D skeletons for action recognition.

There have been several approaches suggested in the recent past to improve recognition performance of actions from such noisy skeletons. These approaches can be mainly divided into two perspectives, namely (i) generative models that assume the skeleton points are produced by a latent dynamic model~\cite{turaga2009locally} corrupted by noise and (ii) discriminative approaches that generate compact representations of sequences on which classifiers are trained~\cite{presti20153d}. Due to the huge configuration space of 3D actions and the unavailability of sufficient training data, discriminative approaches have been the trend in the recent years for this problem. In this line of research, the main idea has been to compactly represent the spatio-temporal evolution of 3D skeletons, and later train classifiers on these representations to recognize the actions. Fortunately, there is a definitive structure to motions of 3D joints relative to each other due to the connectivity and length constraints of body-parts. Such constraints have been used to model actions; examples include Lie Algebra~\cite{vemulapalli_SE3}, positive definite matrices~\cite{harandi2014bregman,hussein2013human}, using a torus manifold~\cite{elgammal2009tracking}, Hanklet representations~\cite{li2012cross}, among several others. While modeling actions with explicit manifold assumptions can be useful, it is computationally expensive. 

In this paper, we present a novel methodology for action representation from 3D skeleton points that avoids any manifold assumptions on the data representation, instead captures the higher-order statistics of how the body-joints relate to each other in a given action sequence. To this end, our scheme combines positive definite kernels and higher-order tensors, with the goal to obtain rich and compact representations. Our scheme benefits from using non-linear kernels such as radial basis functions (RBF) and it can also capture higher-order data statistics and the complexity of action dynamics.

We present two such kernel-tensor representations for the task. Our first representation~\emph{sequence compatibility kernel} (SCK), captures the spatio-temporal compatibility of body-joints between two sequences. To this end, we present an RBF kernel formulation that jointly captures the spatial and temporal similarity of each body-pose (normalized with respect to the hip position) in a sequence against those in another. We show that tensors generated from third-order outer-products of the linearizations of these kernels can be a simple yet powerful representation capturing higher-order co-occurrence statistics of body-parts and yield high classification confidences.

Our second representation, termed~\emph{dynamics compatibility kernel} (DCK) aims at representing spatio-temporal dynamics of each sequence explicitly. 
We present a novel RBF kernel formulation that captures the similarity between a pair of body-poses in a given sequence explicitly, and then compare it against such body-pose pairs in other sequences. As it might appear, such spatio-temporal modeling could be expensive due to the volumetric nature of space and time. However, we show that using an appropriate kernel model can shrink the time-related variable in a small constant size representation after kernel linearization. With this approach, we can model both spatial and temporal variations in the form of co-occurrences which could otherwise have been prohibitive. 

We further show through experiments that the above two representations in fact capture complementary statistics regarding the actions, and combining them leads to significant benefits. We present experiments on three standard datasets for the task, namely (i) UTKinect-Actions~\cite{xia_utkinect}, (ii) Florence3D-Actions~\cite{seidenari_florence3d}, and (iii) MSR-Action3D \cite{li_msraction3d} datasets and demonstrate state-of-the-art accuracy. 

To summarize, the main contributions of this paper are (i) introduction of  sequence and the dynamics compatibility kernels for capturing spatio-temporal evolution of body-joints for 3D skeleton based action sequences, (ii) derivations of linearization of these kernels, and (iii) their tensor reformulations. We review  the related literature next.

\section{Related Work}
\label{sec:related_work}
The problem of skeleton based action recognition has received significant attention over the past decades. Interested readers may refer to useful surveys~\cite{presti20153d} on the topic. In the sequel, we will review some of the more recent related approaches to the problem.

In this paper, we focus on action recognition datasets that represent a human body as an articulated set of connected body-joints that evolve in time~\cite{zatsiorsky_body}. A temporal evolution of the human skeleton is very informative for action recognition as shown by Johansson in his seminal experiment involving the moving lights display~\cite{johansson_lights}. At the simplest level, the human body can be represented as a set of 3D points corresponding to body-joints such as elbow, wrist, knee, ankle, \etc. Action dynamics has been modeled using the motion of such 3D points in~\cite{hussein_action,lv_3daction}, using joint orientations with respect to a reference axis~\cite{parameswaran_viewinvariance} and even relative body-joint positions~\cite{wu_actionlets,yang_eigenjoints}. In contrast, we focus on representing these 3D body-joints by kernels whose linearization results in higher-order tensors capturing complex statistics.  Noteworthy are also parts-based approaches that additionally consider the connected body segments~\cite{yacoob_activities,ohn_hog2,ofli_infjoints,vemulapalli_SE3}.

Our work also differs from previous works in the way it handles the temporal domain. 3D joint locations are modeled as temporal hierarchy of coefficients in \cite{hussein_action}. Pairwise relative positions of joints were modeled in \cite{wu_actionlets} and combined with a hierarchy of Fourier coefficients to capture temporal evolution of actions. Moreover, this approach uses multiple kernel learning to select discriminative joint combinations. In \cite{yang_eigenjoints}, the relative joint positions and their temporal displacements are modeled with respect to the initial frame. In \cite{vemulapalli_SE3}, the displacements and angles between the body parts are represented as a collection of matrices belonging to the special Euclidean group SE(3). Temporal domain is handled by the discrete time warping and Fourier temporal pyramid matching on a sequence of such matrices. In contrast, we model temporal domain with a single RBF kernel providing invariance to local temporal shifts and avoid expensive techniques such as time warping and multiple-kernel learning.
 
Our scheme also differs from prior works such as kernel descriptors \cite{ker_des} that aggregate orientations of gradients for recognition. Their approach exploits sums over the product of at most two RBF kernels handling two cues \eg, gradient orientations and spatial locations, which are later linearized by Kernel PCA and Nystr\"{o}m techniques. Similarly, convolutional kernel networks \cite{ckn} consider stacked layers of a variant of kernel descriptors \cite{ker_des}.
Kernel trick was utilized for action recognition in kernelized covariances \cite{cavazza_kercov} which are obtained in Nystr\"{o}m-like process. A time series kernel \cite{gaidon_timekern} between auto-correlation matrices is proposed to capture spatio-temporal auto-correlations.
In contrast, our scheme allows sums over several multiplicative and additive RBF kernels, thus, it allows handling multiple input cues to build a complex representation. We show how to capture higher-order statistics by linearizing a polynomial kernel and avoid evaluating costly kernels directly in contrast to kernel trick.

Third-order tensors have been found to be useful for several other vision tasks. For example, in \cite{tensoraction2007}, spatio-temporal third-order tensors on videos is proposed for action analysis, non-negative tensor factorization is used for image denoising in~\cite{shashua2005non}, tensor textures are proposed for texture rendering in~\cite{vasilescu2004tensortextures}, and  higher order tensors are used for face recognition in~\cite{vasilescu2002multilinear}. A survey of multi-linear algebraic methods for tensor subspace learning and applications is available in~\cite{lu2011survey}. These applications use a single tensor, while our goal is to use the tensors as data descriptors similar to~\cite{me_tensor2,me_tensor,sparse_tensor_cvpr,zhao2012comprehensive} for image recognition tasks. However, in contrast to these similar methods, we explore the possibility of using third-order representations for 3D action recognition, which poses a different set of challenges. 

\section{Preliminaries}
\label{sec:framework}
In this section, we review our notations and the necessary background on shift-invariant kernels and their linearizations, which will be useful for deriving kernels on 3D skeletons for action recognition.

\subsection{Tensor Notations}
\label{sec:not} 
Let $\tV\in\mbr{d_1\times d_2\times d_3}$ denote a third-order tensor. Using Matlab style notation, we refer to the $p$-th slice of this tensor as $\tV_{:,:,p}$, which is a $d_1\times d_2$ matrix. For a matrix $\mV\in\mbr{d_1\times d_2}$ and a vector $\vv\in\mbr{d_3}$, the notation  $\tV\!=\!\mV\kronstack\vv$ produces a tensor $\tV\!\in\!\mbr{d_1\times d_2\times d_3}$ where the $p$-th slice of $\tV$ is given by $\mV v_p$, $v_p$ being the $p$-th dimension of $\vv$. Symmetric third-order tensors of rank one are formed by the outer product of a vector $\vv\in\mbr{d}$ in modes two and three. That is, a rank-one $\tV\in\mbr{d\times d\times d}$ is obtained from $\vv$ as $\tV\!=\!({\kronstack}_3\vv\!\triangleq\!(\vv\vv^T)\kronstack\vv)$. Concatenation of $n$ tensors in mode $k$ is denoted as $\left[\tV_i\right]_{i\in\idx{n}}^{\oplus_k}$, where $\idx{n}$ is an index sequence $1,2,\cdots, n$. The Frobenius norm of tensor is given by  $\fnorm{\tV} = \sqrt{\sum_{i,j,k} \tVnb_{ijk}^2}$, where $\tVnb_{ijk}$ represents the $ijk$-th element of $\tV$. Similarly, the inner-product between two tensors $\tX$ and $\tY$ is given by $\left\langle\tX,\tY\right\rangle=\sum_{ijk}\tXnb_{ijk}\tYnb_{ijk}$.

\subsection{Kernel Linearization}
\label{sec:kernel_linearization}
Let $G_{\sigma}(\vu-\vub)=\exp(-\enorm{\vu - \vub}^2/{2\sigma^2})$ denote a standard Gaussian RBF kernel centered at $\vub$ and having a bandwidth $\sigma$. Kernel linearization refers to rewriting this $G_{\sigma}$ as an inner-product of two infinite-dimensional feature maps. To obtain these maps, we use a fast approximation method based on probability product kernels \cite{jebara_prodkers}. Specifically, we employ the inner product of $d'$-dimensional isotropic Gaussians given $u,u'\!\!\in\!\mbr{d'}\!$. The resulting approximation can be written as:
%
\begin{align}
&G_{\sigma}\!\left(\vu\!-\!\vub\right)\!\!=\!\!\left(\frac{2}{\pi\sigma^2}\right)^{\!\!\frac{d'}{2}}\!\!\!\!\!\!\int\limits_{\vzeta\in\mbr{d'}}\!\!\!\!G_{\sigma/\sqrt{2}}\!\!\left(\vu\!-\!\vzeta\right)G_{\sigma/\sqrt{2}}(\vub\!\!-\!\vzeta)\,\mathrm{d}\vzeta.
\label{eq:gauss_lin}
\end{align}
Equation \eqref{eq:gauss_lin} is then approximated by replacing the integral with the sum over $Z$ pivots $\vzeta_1,\cdots,\vzeta_Z$, thus writing a feature map $\vphi$ as:
\begin{align}
&\vphi(\vu)=\left[{G}_{\sigma/\sqrt{2}}(\vu-\vzeta_1),\cdots,{G}_{\sigma/\sqrt{2}}(\vu-\vzeta_Z)\right]^T,\!\!\\
\text{ and } & G_{\sigma}(\vu\!-\!\vub)\approx\left<\sqrt{c}\vphi(\vu), \sqrt{c}\vphi(\vub)\right>,
\label{eq:gauss_lin2}
\end{align}
where $c$ represents a constant. We refer to~\eqref{eq:gauss_lin2} as the linearization of the RBF kernel.

\section{Proposed Approach}
In this section, we first formulate the problem of action recognition from 3D skeleton sequences, which precedes an exposition of our two kernel formulations for describing the actions, followed by their tensor reformulations through kernel linearization.

\subsection{Problem Formulation}
Suppose we are given a set of 3D human pose skeleton sequences, each pose consisting of $J$ body-keypoints. Further, to simplify our notations, we assume each sequence consists of $N$ skeletons, one per frame\footnote{\label{foot:foo0}We assume that all sequences have $N$ frames for simplification of presentation. Our formulations are equally applicable to sequences of arbitrary lengths \eg,~$M$ and $N$. Therefore, we apply in practice $G_{\sigma_3}(\frac{s}{M}-\frac{t}{N})$ in Equation \eqref{eq:ker1a}.}. Mathematically, we can define such a pose sequence $\Pi$ as:
\begin{equation}
\Pi = \set{\vx_{is}\in\mbr{3},i\in\idx{J}, s\in\idx{N}}.
\end{equation}
Further, let each such sequence $\Pi$ be associated with one of $K$ action class labels $\ell\in\idx{K}$. Our goal is to use the skeleton sequence $\Pi$ and generate an action descriptor for this sequence that can be used in a classifier for recognizing the action class. In the following, we will present two such action descriptors, namely (i) sequence compatibility kernel and (ii) dynamics compatibility kernel, which are formulated using the ideas of kernel linearization and tensor algebra. We present both these kernel formulations next.

\begin{figure}[t]
\centering
\begin{subfigure}[b]{0.210\linewidth}
\centering\includegraphics[trim=0 0 0 0, clip=true, width=2.6cm]{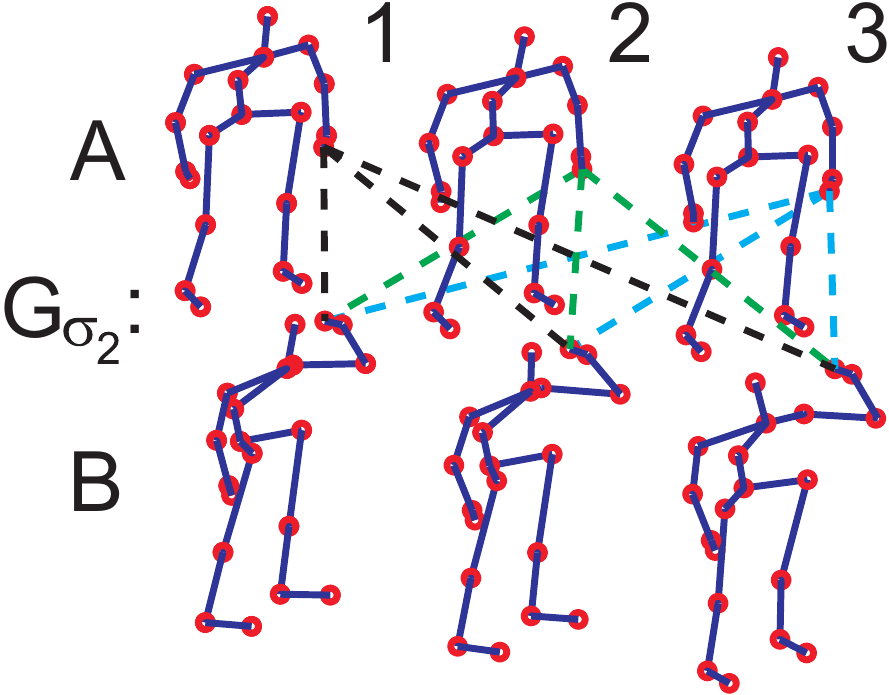}
\caption{\label{fig:ker0a}}
\end{subfigure}
\begin{subfigure}[b]{0.12\linewidth}
\centering\includegraphics[trim=0 0 0 0, clip=true, width=1.42cm]{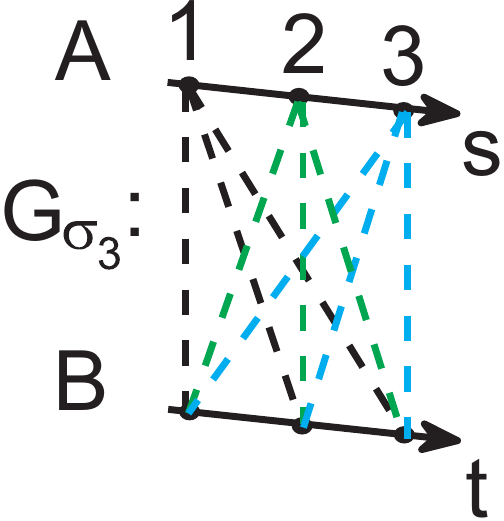}
\caption{\label{fig:ker0a2}}
\end{subfigure}
\begin{subfigure}[b]{0.65\linewidth}
\centering\includegraphics[trim=0 0 0 0, clip=true, width=8.0cm]{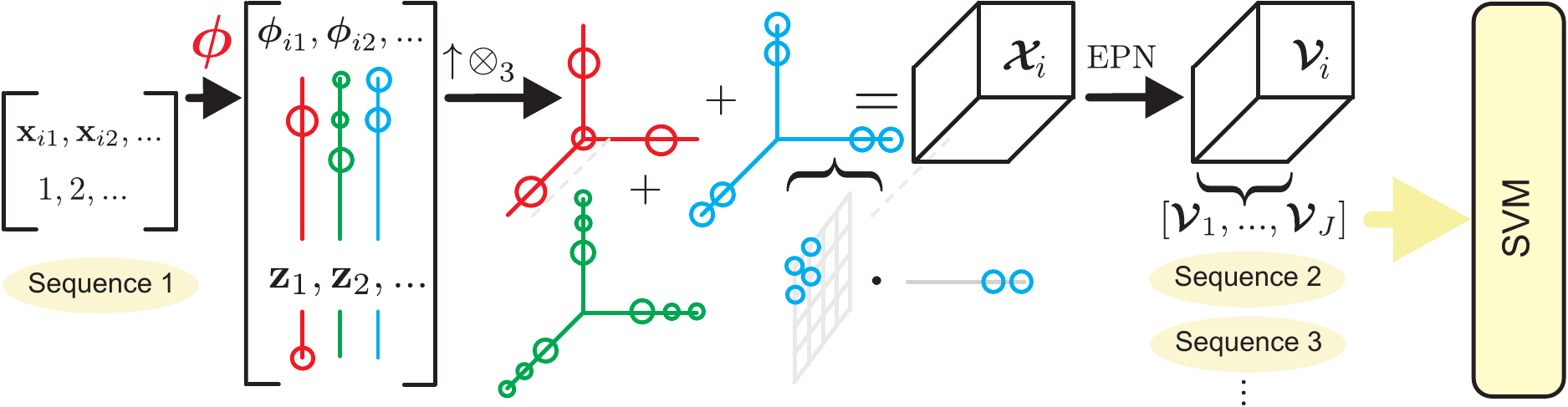}
\caption{\label{fig:ker0b}}
\end{subfigure}
\caption{Figures \ref{fig:ker0a} and \ref{fig:ker0a2} show how SCK works -- kernel $G_{\sigma_2}$ compares exhaustively \eg~hand-related joint $i$ for every frame in sequence $A$ with every frame in sequence $B$. Kernel $G_{\sigma_3}$ compares exhaustively the frame indexes. Figure \ref{fig:ker0b} shows this burden is avoided by linearization -- third-order statistics on feature maps $\vphi(\vx_{is})$ and $\vz(s)$ for joint $i$ are captured in tensor $\tX_i$ and whitened by EPN to obtain $\tV_i$ which are concatenated over $i\!=\!1,\cdots,J$ to represent a sequence.
}
\end{figure}

\subsection{Sequence Compatibility Kernel}
\label{sec:ker1}
As alluded to earlier, the main idea of this kernel is to measure the compatibility between two action sequences in terms of the similarity between their skeletons and their temporal order. To this end, we assume each skeleton is centralized with respect to one of the body-joints (say, hip). Suppose we are given two such sequences $\Pi_A$ and $\Pi_B$, each with $J$ joints, and $N$ frames. Further, let $\vx_{is}\!\in\!\mbr{3}$ and $\vy_{jt}\!\in\!\mbr{3}$ correspond to the  body-joint coordinates of $\Pi_A$ and $\Pi_B$, respectively. We define our~\emph{sequence compatibility kernel} (SCK) between $\Pi_A$ and $\Pi_B$ as\footref{foot:foo0}:
\begin{align}
& K_S(\Pi_A,\Pi_B) = \frac{1}{\Lambda}\!\!\! \sum\limits_{(i,s)\in\idxJ}\sum\limits_{(j,t)\in\idxJ}\!G_{\sigma_1}(i\!-\!j)\Big(\beta_1 G_{\sigma_2}\!\left(\vx_{is} - \vy_{jt}\right) + \beta_2\, G_{\sigma_3}(\frac{s-t}{N})\Big)^r,\label{eq:ker1a}
\end{align}
where $\Lambda$ is a normalization constant and $\idxJ=\idx{J}\times\idx{N}$. As is clear, this kernel involves three different compatibility subkernels, namely (i) $G_{\sigma_1}$, that captures the compatibility between joint-types $i$ and $j$, (ii) $G_{\sigma_2}$, capturing the compatibility between joint locations $\vx$ and $\vy$, and (iii) $G_{\sigma_3}$, measuring the temporal alignment of two poses in the sequences. We also introduce weighting factors $\beta_1,\beta_2\geq 0$ that adjusts the importance of the body-joint compatibility against the temporal alignment, where $\beta_1+\beta_2=1$. Figures \ref{fig:ker0a} and \ref{fig:ker0a2} illustrate how this kernel works. It might come as a surprise, why we need the kernel $G_{\sigma_1}$. Note that our skeletons may be noisy and there is a possibility that some of the keypoints are detected incorrectly (for example, elbows and wrists). Thus, this kernel allows incorporating some degree of uncertainty to the alignment of such joints. To simplify our formulations, in this paper, we will assume that such errors are absent from our skeletons, and thus $G_{\sigma_1}(i-j)=\delta(i-j)$. Further, the standard deviations $\sigma_2$ and $\sigma_3$ control the joint-coordinate selectivity and temporal shift-invariance respectively. That is, for $\sigma_3\rightarrow 0$, two sequences will have to match perfectly in the temporal sense. For $\sigma_3\rightarrow\infty$, the algorithm is invariant to any permutations of the frames. As will be clear in the sequel, the parameter $r$ determines the order statistics of the kernel (we use $r=3$).

Next, we present linearization of our kernel using the method proposed in Section~\ref{sec:kernel_linearization} and Equation~\eqref{eq:gauss_lin2} 
so that kernel $G_{\sigma_2}(\vx-\vy)\approx \phi(\vx)^T\phi(\vy)$ (see note\footnote{\label{foot:foob}In practice, we use $G^{'}_{\sigma_2}(\vx\!-\!\vy)\!=\!G_{\sigma_2}(x^{(x)}\!\!-\!y^{(x)})\!+\!G_{\sigma_2}(x^{(y)}\!\!-\!y^{(y)})\!+\!G_{\sigma_2}(x^{(z)}\!\!-\!y^{(z)})$ so the kernel $G^{'}_{\sigma_2}(\vx\!-\!\vy)\approx[\phi(x^{(x)}\!); \phi(x^{(y)}\!); \phi(x^{(z)}\!)]^T\![\phi(y^{(x)}\!); \phi(y^{(y)}\!); \phi(y^{(z)}\!)]$ but for simplicity we write $G_{\sigma_2}(\vx\!-\!\vy)\!\approx\!\phi(\vx)^T\phi(\vy)$. Note that $(x)$, $(y)$, $(z)$ are the spatial xyz-components of joints.}) while $G_{\sigma_3}(\frac{s-t}{N})\approx\vz(s/N)^T\vz(t/N)$. With these approximations and simplification to $G_{\sigma_1}\!$ we described above, we can rewrite our sequence compatibility kernel as:
\begin{align}
K_S(\Pi_A,\Pi_B) &= \!\frac{1}{\Lambda}\!\!\sum\limits_{i\in\idx{J}}\sum\limits_{s\in\idx{N}}\!\sum\limits_{t\in\idx{N}}\!\!\!\left(
\begin{bmatrix}
\sqrt{\beta_1}\,\vphi(\vx_{is}),\text{ (see note\footref{foot:foob})}\\
\sqrt{\beta_2}\,\vz(s/N)\\[3pt]
\end{bmatrix}^T\!\!\!\cdot
\begin{bmatrix}
\sqrt{\beta_1}\vphi(\vy_{it})\\
\sqrt{\beta_2}\vz(t/N)\\[3pt]
\end{bmatrix}\right)^r\\
&=\!\frac{1}{\Lambda}\!\!\sum\limits_{i\in\idx{J}}\sum\limits_{s\in\idx{N}}\!\sum\limits_{t\in\idx{N}}\!\!\!\left<
{\kronstack}_r\!\begin{bmatrix}
\sqrt{\beta_1}\,\vphi(\vx_{is})\\
\sqrt{\beta_2}\,\vz(s/N)\\[3pt]
\end{bmatrix}\!,
{\kronstack}_r\!\begin{bmatrix}
\sqrt{\beta_1}\vphi(\vy_{it})\\
\sqrt{\beta_2}\vz(t/N)\\[3pt]
\end{bmatrix}\right>\\
&=\!\!\!\sum\limits_{i\in\idx{J}}\!\!\left<
\!\frac{1}{\sqrt{\Lambda}}\!\!\sum\limits_{s\in\idx{N}}\!\!{\kronstack}_r\!\begin{bmatrix}
\sqrt{\beta_1}\,\vphi(\vx_{is})\\
\sqrt{\beta_2}\vz(s/N)\\[3pt]
\end{bmatrix}\!,
\frac{1}{\sqrt{\Lambda}}\!\!\sum\limits_{t\in\idx{N}}\!\!{\kronstack}_r\!\begin{bmatrix}
\sqrt{\beta_1}\vphi(\vy_{it})\\
\sqrt{\beta_2}\vz(t/N)\\[3pt]
\end{bmatrix}\right>.
\label{eq:ker1b}
\end{align}

As is clear,~\eqref{eq:ker1b} expresses $K_S(\Pi_A,\Pi_B)$ as a sum of inner-products on third-order tensors ($r=3$). This is illustrated by Figure \ref{fig:ker0b}. While, using the dot-product as the inner-product is a possibility, there are much richer alternatives for tensors of order $r>=2$ that can exploit their structure or manipulate higher-order statistics inherent in them, thus leading to better representations. An example of such a commonly encountered property is the so-called \emph{burstiness}~\cite{jegou_bursts}, which is the property that a given feature appears more often in a sequence than a statistically independent model would predict. A robust sequence representation should be invariant to the length of actions \eg, a prolonged \emph{hand waving} represents the same action as a short \emph{hand wave}. The same is true for short versus repeated \emph{head nodding}. Eigenvalue Power Normalization (EPN)~\cite{me_tensor} is known to suppress burstiness. It acts on higher-order statistics illustrated in Figure~\ref{fig:ker0b}. Incorporating EPN, we generalize~\eqref{eq:ker1b} as:
\begin{align}
&
\!K_S^{*}(\piA,\piB)\!=\!\!\!\sum\limits_{i\in\idx{J}}\!\!\left<
\!\mygthree{\frac{1}{\sqrt{\Lambda}}\!\!\sum\limits_{s\in\idx{N}}\!\!\!{\kronstack}_r\!\!\begin{bmatrix}
\!\sqrt{\beta_1}\vphi(\vx_{is})\\
\sqrt{\beta_2}\vz(s/N)\\[3pt]
\end{bmatrix}}\!\!,\mygthree{\frac{1}{\sqrt{\Lambda}}\!\!\sum\limits_{t\in\idx{N}}\!\!\!{\kronstack}_r\!\!\begin{bmatrix}
\sqrt{\beta_1}\vphi(\vy_{it})\\
\sqrt{\beta_2}\vz(t/N)\\[3pt]
\end{bmatrix}}\!\!\right>\!,\label{eq:ker1c}
\end{align}
where the operator $\tG$ performs EPN by applying power normalization to the spectrum of the third-order tensor (by taking the higher-order SVD). Note that in general $K_S^{*}(\piA,\piB)\!\not\approx\!K_S(\piA,\piB)$ as $\tG$ is intended to manipulate the spectrum of $\tX$. The final representation, for instance for a sequence $\piA$, takes the following form:
\begin{align}
& \tV_i\!=\!\mygthree{\tX_i}\!,\text{ where } \tX_i\!=\!\!\frac{1}{\sqrt{\Lambda}}\!\!\!\sum\limits_{s\in\idx{N}}\!\!\!{\kronstack}_r\!\!\begin{bmatrix}
\!\sqrt{\beta_1}\,\vphi(\vx_{is})\\
\sqrt{\beta_2}\vz(s/N)\\[3pt]
\end{bmatrix}\!.\label{eq:ker1d}
\end{align}
We can further replace the summation over the body-joint indexes in \eqref{eq:ker1c} by concatenating $\tV_i$ in ~\eqref{eq:ker1d} along the fourth tensor mode, thus defining $\tV = \big[\tV_i\big]_{i\in\idx{J}}^{\oplus_4}$. Suppose $\tV_A$ and $\tV_B$ are the corresponding fourth order tensors for $\Pi_A$ and $\Pi_B$ respectively. 
Then, we obtain:
\begin{align}
& K_S^{*}(\piA,\piB)=\left<\tV_A, \tV_B\right>.
\end{align}

Note that the tensors $\tX$ have the following properties: (i) super-symmetry $\tX_{i,j,k}\!=\!\tX_{\pi(i,j,k)}$ for indexes $i,j,k$ and their permutation given by $\pi,\;\forall\pi$, and (ii) positive semi-definiteness of every slice, that is, $\tX_{:,:,s}\!\in\!\semipd{d},$ for $s\!\in\!\idx{d}$. Therefore, we need to use only the upper-simplex of the tensor which consists of $\binom{d+r-1}{r}$ coefficients (which is the total size of our final representation) rather than $d^r\!$, where $d$ is the side-dimension of $\tX$ \ie, $d\!=\!3Z_2\!+\!Z_3$ (see note\footref{foot:foob}), and $Z_2$ and $Z_3$ are the numbers of pivots used in the approximation of $G_{\sigma_2}$ (see note\footref{foot:foob}) and $G_{\sigma_3}$ respectively. As we want to preserve the above listed properties in tensors $\tV$, we employ slice-wise EPN which is induced by the Power-Euclidean distance and involves rising matrices to a power $\gamma$. Finally, we re-stack these slices along the third mode as:
\begin{align}
& \mygthree{\tX}\!=\![\tX_{:,:,s}^{\gamma}]_{s\in\idx{d}}^{\oplus_3}, \text{ for } 0\!<\gamma\!\leq\!1.\label{eq:epn1}
\end{align}
This $ \mygthree{\tX}$ forms our tensor representation for the action sequence.

\begin{figure}[t]
\centering
\begin{subfigure}[b]{0.195\linewidth}
\centering
\includegraphics[trim=0 0 0 0, clip=true, width=2.95cm]{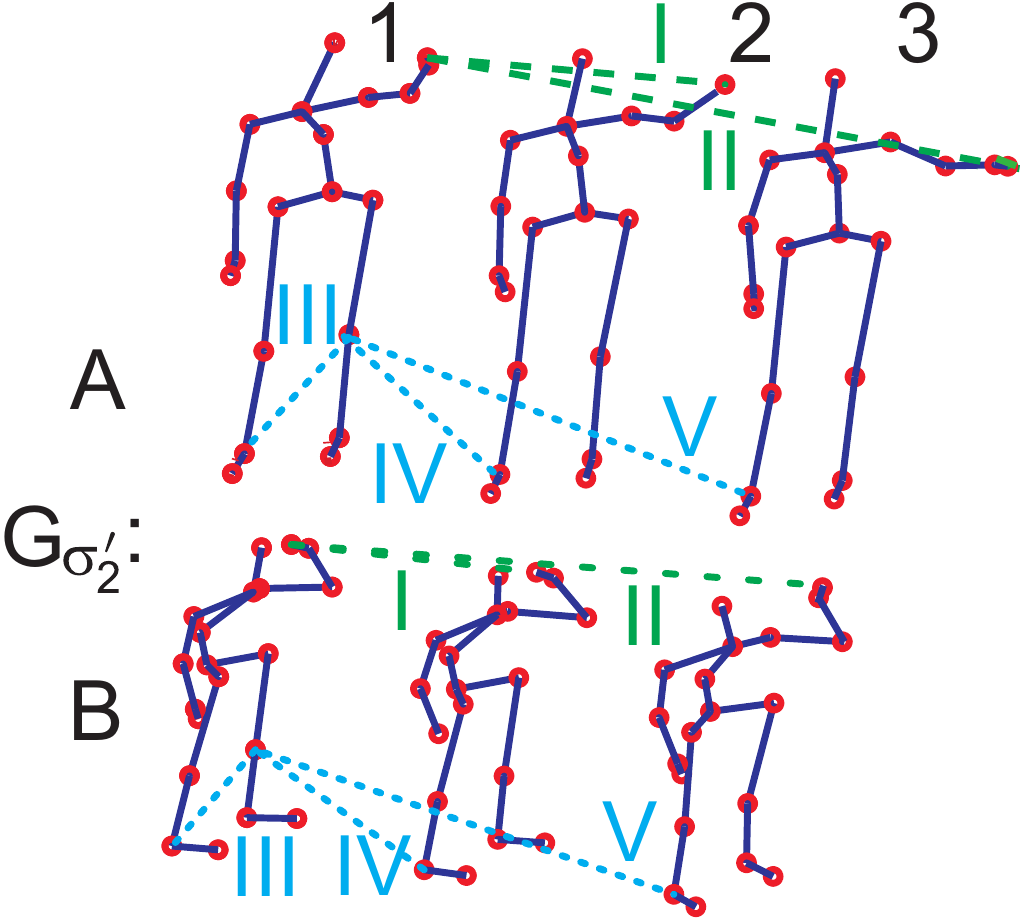}
\caption{\label{fig:ker2a}}
\end{subfigure}
\begin{subfigure}[b]{0.135\linewidth}
\centering
\includegraphics[trim=0 0 0 0, clip=true, width=1.6cm]{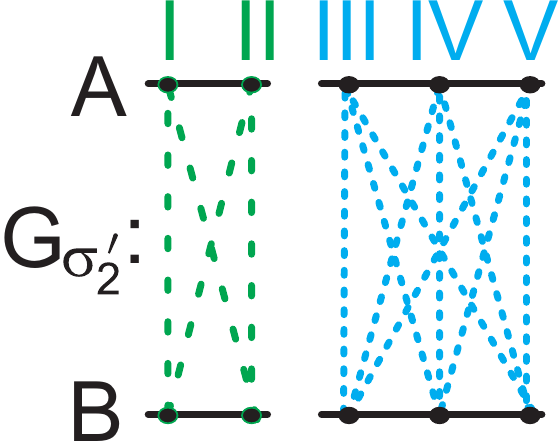}
\caption{\label{fig:ker2b}}
\end{subfigure}
\begin{subfigure}[b]{0.65\textwidth}
\centering
\includegraphics[trim=0 0 0 0, clip=true, width=8.0cm]{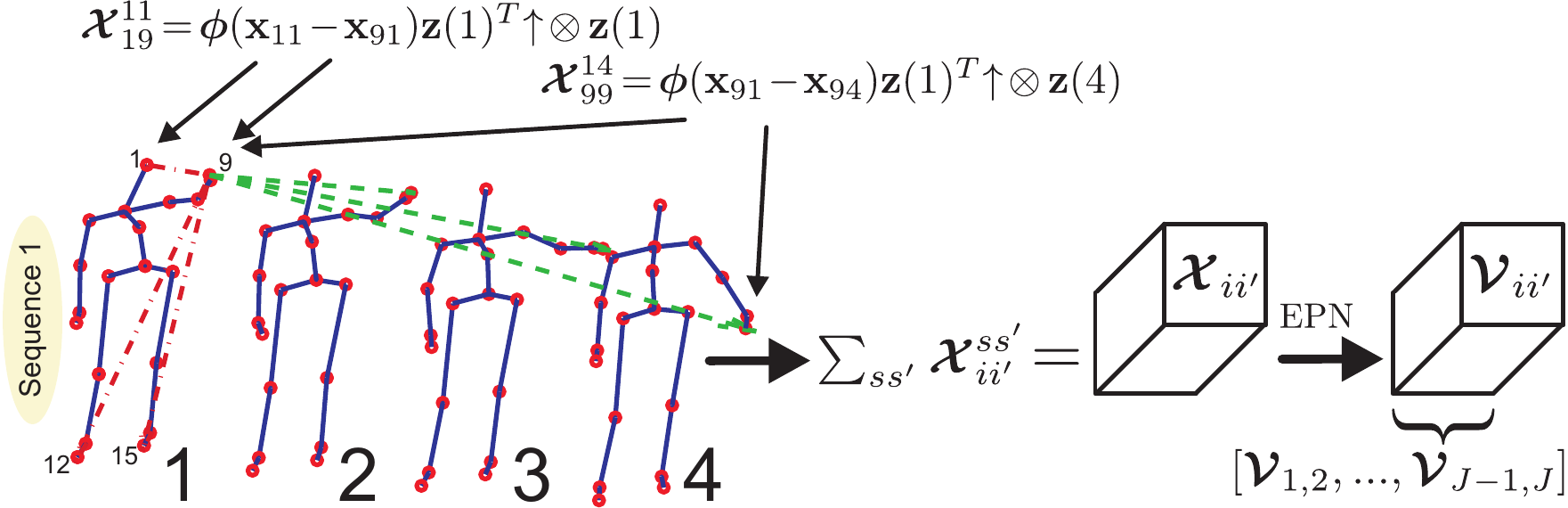}
\caption{\label{fig:ker2c}}
\end{subfigure}
\caption{Figure \ref{fig:ker2a} shows that kernel $G_{\sigma'_2}$ in DCK captures spatio-temporal dynamics by measuring displacement vectors from any given body-joint to remaining joints spatially- and temporally-wise (\ie~see dashed lines). Figure \ref{fig:ker2b} shows that comparisons performed by $G_{\sigma'_2}$ for any selected two joints are performed all-against-all temporally-wise which is computationally expensive. Figure \ref{fig:ker2c} shows the encoding steps in the proposed linearization which overcome this burden.}
\end{figure}

\subsection{Dynamics Compatibility Kernel}
The SCK kernel that we described above captures the inter-sequence alignment, while the intra-sequence spatio-temporal dynamics is lost. In order to capture these temporal dynamics, we propose a novel dynamics compatibility kernel (DCK). To this end, we use the absolute coordinates of the joints in our kernel. Using the notations from the earlier section, for two action sequences $\Pi_A$ and $\Pi_B$, we define this kernel as:
\begin{align}
& \!\!\!K_D(\piA,\piB)=\frac{1}{\Lambda}\!\!\!\!\!\sum\limits_{\substack{(i,s)\in\idxJ\!,\\(i',s')\in\idxJ\!,\\i'\!\!\neq\!i\!,s'\!\!\neq\!s}}\sum\limits_{\substack{(\!j,t)\in\idxJ\!,\\(\!j'\!\!,t'\!)\in\idxJ,\\j'\!\!\neq\!j\!,t'\!\!\neq\!t}}\!\!\!\!G'_{\sigma'_1}(i\!-\!j\!, i'\!\!-\!j'\!)\,G_{\sigma'_2}\left(\left(\vx_{is}\!-\!\vx_{i's'}\!\right)\!-\!\left(\vy_{jt}\!-\!\vy_{j't'}\right)\right)\cdot\nonumber\\[-16pt]
& \qquad\qquad\qquad\qquad\qquad\qquad\qquad\qquad\quad\cdot G'_{\sigma'_3}(\frac{s\!-\!t}{N},\!\frac{s'\!\!-\!t'}{N})\,G'_{\sigma'_4}(s\!-\!s'\!,t\!-\!t'\!),\label{eq:ker2a}
\end{align}
where $G'_{\sigma}(\valpha,\vbeta)=G_{\sigma}(\valpha)G_{\sigma}(\vbeta)$. In comparison to the SCK kernel in~\eqref{eq:ker1a}, the DCK kernel uses the intra-sequence joint differences, thus capturing the dynamics. This dynamics is then compared to those in the other sequences. Figures~\ref{fig:ker2a}-\ref{fig:ker2c} depict schematically how this kernel captures co-occurrences. As in SCK, the first kernel $G'_{\sigma'_1}$ is used to capture sensor uncertainty in body-keypoint detection, and is assumed to be a delta function in this paper. The second kernel $G_{\sigma'_2}$ models the spatio-temporal co-occurrences of the body-joints. Temporal alignment kernels expressed as $G_{\sigma'_3}$ encode the temporal start and end-points from $(s,s'\!)$ and $(t,t'\!)$. Finally, $G_{\sigma'_4}$ limits contributions of dynamics between temporal points if they are distant from each other, \ie~if $s'\!\gg\!s$ or $t'\!\gg\!t$ and $\sigma'_4$ is small. Furthermore, similar to SCK, the standard deviations $\sigma'_2$ and $\sigma'_3$ control the selectivity over spatio-temporal dynamics of body-joints and their temporal shift-invariance for the start and end points, respectively.. As discussed for SCK, the practical extensions described by the footnotes\footref{foot:foo0}\textsuperscript{,}\footref{foot:foob} apply to DCK as well. 

As in the previous section, we employ linearization to this kernel. Following the derivations described above, it can be shown that the linearized kernel has the following form (see appendices for details):
\begin{align}
&\!K_D(\Pi_A,\Pi_B) =\!\!\!\!\sum\limits_{\substack{i\in\idx{J}\!,\\i'\!\in\idx{J}\!:\\i'\!\neq i}}
\!\!
\left<
\!\frac{1}{\sqrt{\Lambda}}\!\!\sum\limits_{\substack{s\in\idx{N}\!,\\s'\!\!\in\idx{N}\!:\\s'\!\!\neq\!s}}\!\!
G_{\sigma'_4}(s\!-\!s'\!)\left(\vphi(\vx_{is}\!\!-\!\!\vx_{i's'})
\!\cdot\!\vz\big(\frac{s}{N}\big)^T\!\right)\!\kronstack\vz\big(\frac{s'\!}{N}\big)\!\right.,\label{eq:ker2b}\\[-16pt]
&\qquad\qquad\qquad\qquad\qquad\qquad\;
\left.
\!\frac{1}{\sqrt{\Lambda}}\!\!\sum\limits_{\substack{t\in\idx{N}\!,\\t'\!\!\in\idx{N}\!:\\t'\!\!\neq\!t}}\!\!
G_{\sigma'_4}(t\!-\!t'\!)\Big(
\vphi(\vy_{it}\!\!-\!\!\vy_{i't'})
\!\cdot\!\vz\big(\frac{t}{N}\big)^T\!\Big)\!\kronstack\vz\big(\frac{t'\!}{N}\big)\!\right>\!.\nonumber
\end{align}

Equation~\eqref{eq:ker2b} expresses $K_D(\piA,\piB)$ as a sum over inner-products on third-order non-symmetric tensors of third-order (c.f. Section \ref{sec:ker1}  where the proposed kernel results in an inner-product between super-symmetric tensors). However, we can decompose each of these tensors with a variant of EPN which involves Higher Order Singular Value Decomposition (HOSVD) into factors stored in the so-called core tensor and equalize the contributions of these factors. Intuitively, this would prevent bursts in the statistically captured spatio-temporal co-occurrence dynamics of actions. For example, consider that a long \emph{hand wave} versus a short one yield different temporal statistics, that is, the prolonged action results in bursts. However, the representation for action recognition should be invariant to such cases. As in the previous section, we introduce a non-linear operator $\tG$ into equation \eqref{eq:ker2b} which will handle this. Our final representation, for example, for sequence $\piA$ can be expressed as:
\begin{align}
& \!\!\!\!\!\tV_{ii'\!}\!=\!\mygthree{\tX_{ii'\!}}\!,\!\text{ and }\tX_{ii'\!}\!=\!\!\frac{1}{\sqrt{\Lambda}}\!\!\sum\limits_{\substack{s\in\idx{N}\!,\\s'\!\!\in\idx{N}\!:\\s'\!\!\neq\!s}}\!\!
G_{\sigma'_4}(s\!-\!s'\!)\left(
\vphi(\vx_{is}\!\!-\!\!\vx_{i's'})
\!\cdot\!\vz\big(\frac{s}{N}\big)^T\!\right)\!\kronstack\vz\big(\frac{s'\!}{N}\big),\!\!\label{eq:ker2d}
\end{align}
where the summation over the pairs of body-joint indexes in \eqref{eq:ker2b} becomes equivalent to the concatenation of $\tV_{ii'}\!$ from \eqref{eq:ker2d} along the fourth mode such that we obtain tensor representations $\big[\tV_{ii'\!}\big]_{i>i'\!:\,i,i'\in\idx{J}}^{\oplus_4}\!$  for sequence $\piA$ and $\big[\tVH_{ii'\!}\big]_{i>i'\!:\,i,i'\in\idx{J}}^{\oplus_4}\!$ for sequence $\piB$. The dot-product can be now applied between these representations for comparing them. For the operator $\tG$, we choose HOSVD-based tensor whitening as proposed in \cite{me_tensor}. However, they work with the super-symmetric tensors, such as the one we proposed in Section \ref{sec:ker1}. We work with a general non-symmetric case in \eqref{eq:ker2d} and use the following operator $\tG$:
\begin{align}
&{\left(\tE; \vec{A}_1,\cdots,\vec{A}_r\right)}=\hosvd(\tX)\label{eq:rawcod3}\\
&\tEH=\sgn\!\left(\tE\right)\!\,\left|\!\,\tE\right|^{\gamma}\label{eq:rawcod4}\\
&\tVT=((\tEH\otimes_{1}\!\vec{A}_1)\,\cdots)\otimes_{r}\!\vec{A}_r\label{eq:rawcod5}\\
&\tG(\tX)=\sgn(\tVT)\,|\!\tVT|^{\gamma^{*}}\label{eq:rawcod6}
\end{align}
In the above equations, we distinguish the core tensor $\tE$ and its power normalized variants $\tEH$ with factors that are being evened out by rising to the power $0\!<\!\gamma\!\leq\!1$, eigenvalue matrices $\vec{A}_1,\cdots,\vec{A}_r$ and operation $\otimes_r$ which represents a so-called tensor-product in mode $r$. We refer the reader to paper \cite{me_tensor} for the detailed description of the above steps.

\section{Computational Complexity}
Non-linearized SCK with kernel SVM has complexity $\bigoh(JN^2T^\rho)$ given $J$ body joints, $N$ frames per sequence, $T$ sequences, and $2\!<\!\rho\!<\!3$ which concerns complexity of kernel SVM. Linearized SCK with linear SVM takes $\bigoh(JNTZ_*^r)$ for a total of $Z_*$ pivots and tensor order $r\!=\!3$. Note that $N^2T^\rho\!\gg\!NTZ_*^r$. For $N\!=\!50$ and $Z_*\!=\!20$,  this is $3.5\!\times$ (or $32\!\times$) faster than the exact kernel for $T\!=\!557$ (or $T\!=\!5000$) used in our experiments. Non-linearized DCK with kernel SVM has complexity $\bigoh(J^2N^4T^\rho)$ while linearized DCK takes $\bigoh(J^2N^2TZ^3)$ for $Z$ pivots per kernel, \eg~$Z\!=\!Z_2\!=\!Z_3$ given $G_{\sigma'_2}$ and $G_{\sigma'_3}$. As $N^4T^\rho\!\gg\!N^2TZ^3$, the linearization is $~11000\!\times$ faster than the exact kernel, for say $Z\!=\!5$. Note that EPN incurs negligible cost (see appendices for details).

\section{Experiments}\label{sec:exp}
In this section, we present experiments using our models on three benchmark 3D skeleton based action recognition datasets, namely (i) the UTKinect-Action~\cite{xia_utkinect}, (ii) Florence3D-Action~\cite{seidenari_florence3d}, and (iii) MSR-Action3D~\cite{li_msraction3d}. We also present experiments evaluating the influence of the choice of various hyper-parameters, such as the number of pivots $Z$ used for linearizing the body-joint and temporal kernels, the impact of Eigenvalue Power Normalization, and factor equalization.

\subsection{Datasets}
\label{sec:sets}
\noindent\textbf{UTKinect-Action~\cite{xia_utkinect}} dataset consists of 10 actions performed twice by 10 different subjects, and has 199 action sequences. The dataset provides 3D coordinate annotations of 20 body-joints for every frame. The dataset was captured with a stationary Kinect sensor and contains significant viewpoint and intra-class variations.
\\
\noindent\textbf{Florence3D-Action~\cite{seidenari_florence3d}} dataset consists of 9 actions performed two to three times by 10 different subjects. It comprises 215 action sequences. 3D coordinate annotations of 15 body-joints are provided for every frame. This dataset was also captured with a Kinect sensor and contains significant intra-class variations \ie, the same action may be articulated with the left or right hand. Moreover, some actions such as \emph{drinking}, \emph{performing a phone call}, etc., can be visually ambiguous.
\\
\noindent\textbf{MSR-Action3D~\cite{li_msraction3d}} dataset is comprised from 20 actions performed two to three times by 10 different subjects. Overall, it consists of 557 action sequences. 3D coordinates of 20 body-joints are provided. This dataset was captured using a Kinect-like depth sensor. It exhibits strong inter-class similarity.

In all experiments we follow the standard protocols for these datasets. We use the cross-subject test setting, in which half of the subjects  are used for training and the remaining half for testing. Similarly, we divide the training set into two halves for purpose of training-validation. Additionally, we use two protocols for MSR-Action3D according to approaches~\cite{wu_actionlets} and ~\cite{li_msraction3d}, where the latter protocol uses three subsets grouping related actions together.

\subsection{Experimental Setup}\label{sec:setup}

For the sequence compatibility kernel, we first normalized all body-keypoints with respect to the hip joints across frames, as indicated in Section \ref{sec:ker1}. Moreover, lengths of all body-parts are normalized with respect to a reference skeleton. This setup follows the pre-processing suggested in~\cite{vemulapalli_SE3}. For our dynamics compatibility kernel, we use unnormalized body-joints and assume that the displacements of body-joint coordinates across frames capture their temporal evolution implicitly.

\noindent{\textbf{Sequence compatibility kernel.}} In this section, we first present experiments evaluating the influence of parameters $\sigma_2$ and $\sigma_3$ of kernels $G_{\sigma_2}$ and $G_{\sigma_3}$ which control the degree of selectivity for the 3D body-joints and temporal shift invariance, respectively. See Section \ref{sec:ker1} for a full definition of these parameters.

\begin{figure}[t]
\centering\hspace*{-0.1cm}
\begin{subfigure}[b]{0.32\linewidth}
\centering\includegraphics[trim=0 3 0 15, clip=true, width=4.35cm]{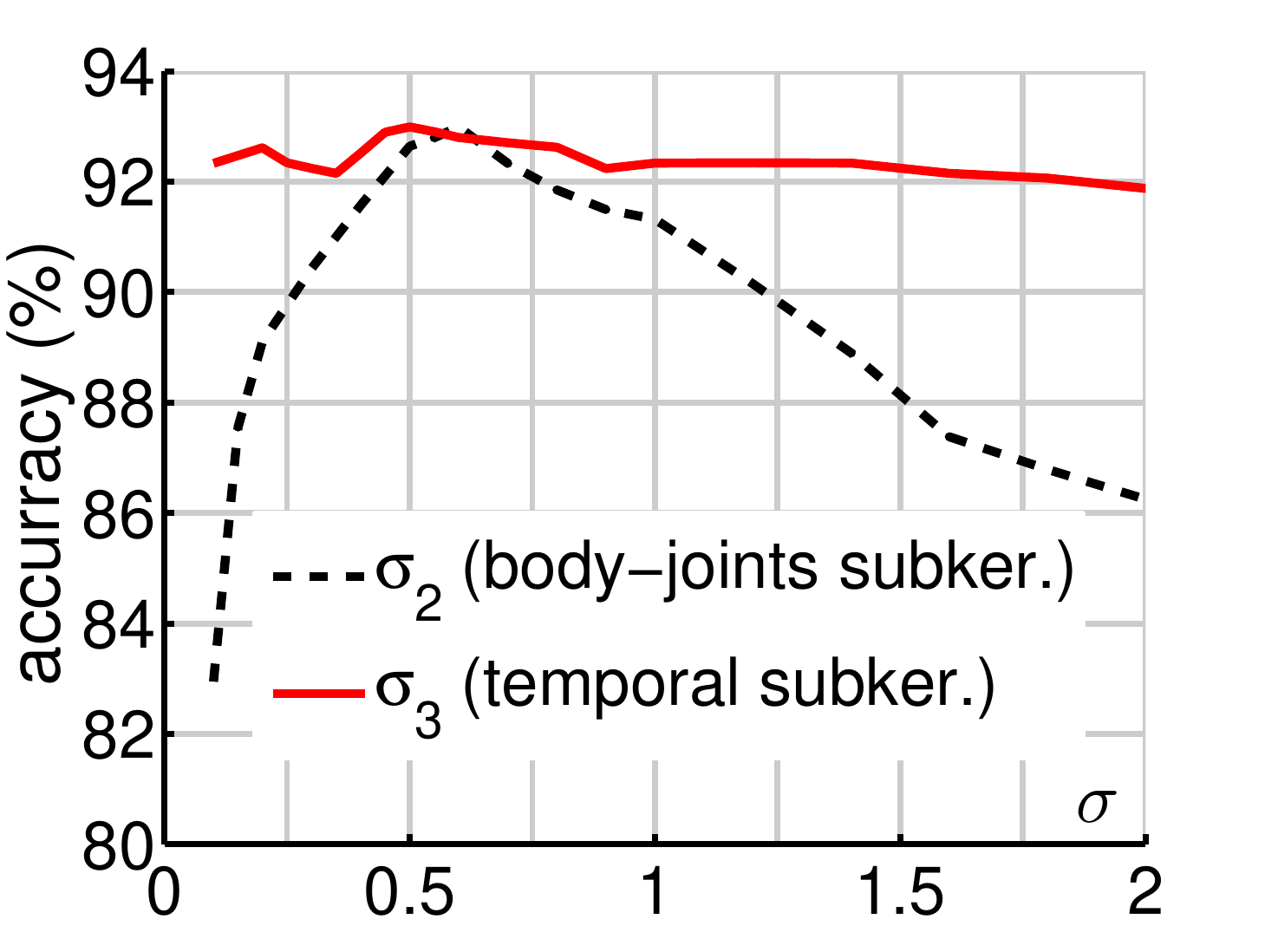}
\caption{\label{fig:piv1}}
\end{subfigure}
\hspace*{0.05cm}
\begin{subfigure}[b]{0.32\linewidth}
\centering\includegraphics[trim=0 3 0 15, clip=true, width=4.35cm]{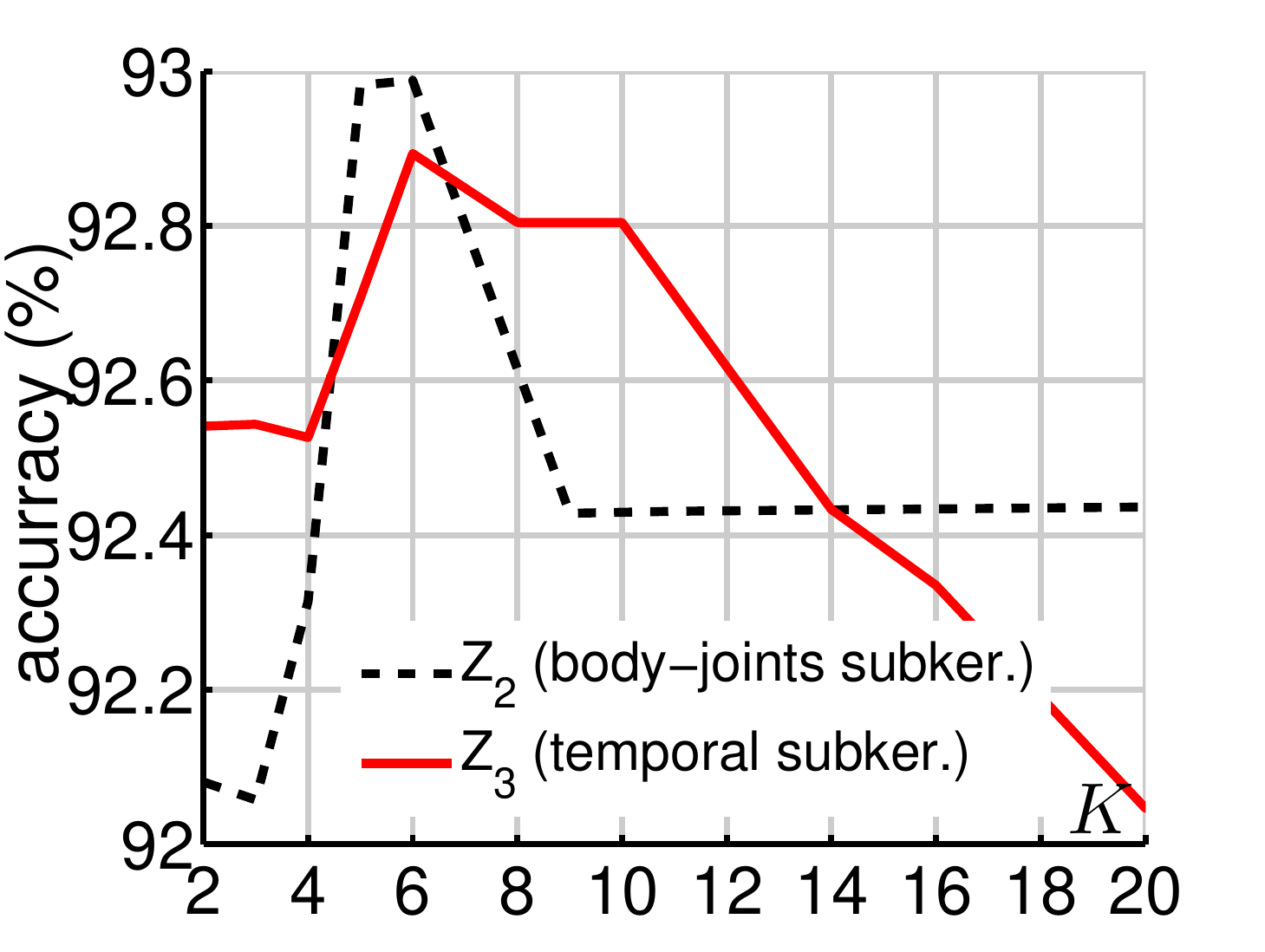}
\caption{\label{fig:piv2}}
\end{subfigure}
\hspace*{0.05cm}
\begin{subfigure}[b]{0.32\linewidth}
\includegraphics[trim=0 3 0 15, clip=true, width=4.35cm]{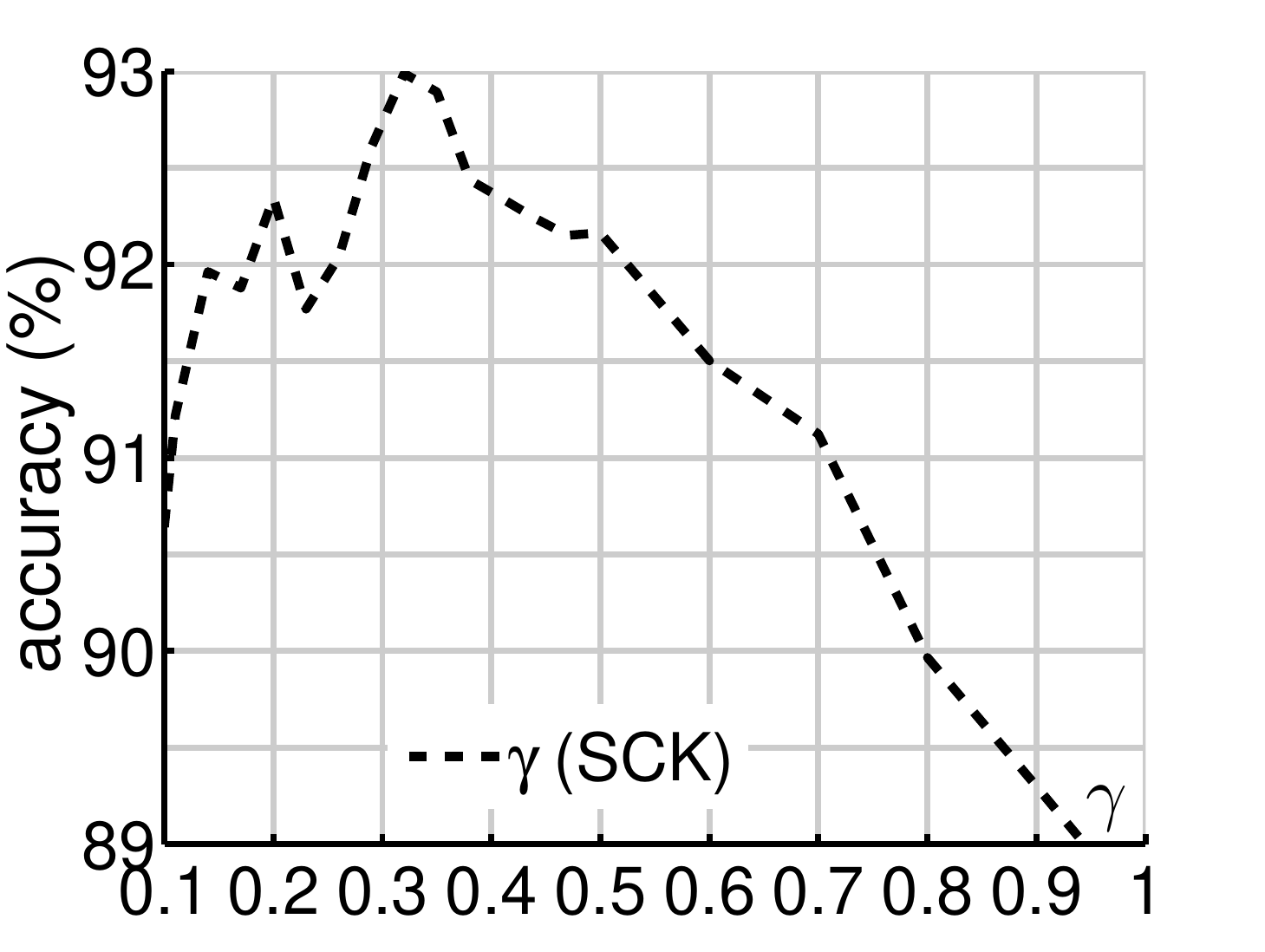}
\caption{\label{fig:piv3}}
\end{subfigure}
\caption{Figure \ref{fig:piv1} illustrates the classification accuracy on Florence3d-Action for the sequence compatibility kernel when varying radii $\sigma_2$ (body-joints subkernel) and $\sigma_3$ (temporal subkernel). Figure \ref{fig:piv2} evaluates behavior of SCK w.r.t. the number of pivots $Z_2$ and $Z_3$. Figure \ref{fig:piv3} demonstrates effectiveness of our slice-wise Eigenvalue Power Normalization in tackling burstiness by varying parameter $\gamma$.}
\end{figure}

Furthermore, recall that the kernels $G_{\sigma_2}$ and $G_{\sigma_3}$ are approximated via linearizations according to equations \eqref{eq:gauss_lin} and \eqref{eq:gauss_lin2}. The quality of these approximations and the size of our final tensor representations depend on the number of pivots $Z_2\!$ and $Z_3$ chosen. In our experiments, the pivots $\vzeta$ are spaced uniformly within interval $[-1;1]$ and $[0;1]$ for kernels $G_{\sigma_2}$ and $G_{\sigma_3}$ respectively.

Figures \ref{fig:piv1} and \ref{fig:piv2} present the results of this experiment on the Florence3D-Action dataset -- these are the results presented on the test set as we have also observed exactly the same trends on the validation set.

Figure \ref{fig:piv1} shows that the body-joint compatibility subkernel $G_{\sigma_2}$ requires a choice of $\sigma_2$ which is not too strict as the specific body-joints (\eg, elbow) would be expected to repeat across sequences in the exactly same position. On the one hand, very small $\sigma_2$ leads to poor generalization. On the other hand, very large $\sigma_2$ allows big displacements of the corresponding body-joints between sequences which results in poor discriminative power of this kernel. Furthermore, Figure \ref{fig:piv1} demonstrates that the range of $\sigma_3$ for the temporal subkernel for which we obtain very good performance is large, however, as $\sigma_3$ becomes very small or very large, extreme temporal selectivity or full temporal invariance, respectively, result in a loss of performance. For instance, $\sigma_3\!=\!4$ results in $91\%$ accuracy only.

In Figure \ref{fig:piv2}, we show the performance of our SCK kernel with respect to the number of pivots used for linearization. For the body-joint compatibility subkernel $G_{\sigma_2}$, we see that $Z_2\!=\!5$ pivots are sufficient to obtain good performance of $92.98\%$ accuracy. We have observed that this is consistent with the results on the validation set. Using more pivots, say $Z_2\!=\!20$, deteriorates the results slightly, suggesting overfitting. We make similar observations for the temporal subkernel $G_{\sigma_3}$ which demonstrates good performance for as few as $Z_3\!=\!2$ pivots. Such a small number of pivots suggests that linearizing 1D variables and generating higher-order co-occurrences, as described in Section~\ref{sec:ker1}, is a simple, robust, and effective linearization strategy.

Further, Figure \ref{fig:piv3} demonstrates the effectiveness of our slice-wise Eigenvalue Power Normalization (EPN) described in Equation \eqref{eq:epn1}. When $\gamma\!=\!1$, the EPN functionality is absent. This results in a drop of performance from $92.98\%$ to $88.7\%$ accuracy. This demonstrates that statistically unpredictable bursts of actions described by the body-joints, such as long versus short \emph{hand waving}, are indeed undesirable. It is clear that in such cases, EPN is very effective, as in practice it considers correlated bursts, \eg~co-occurring \emph{hand wave} and associated with it elbow and neck motion. For more details behind this concept, see~\cite{me_tensor}. For our further experiments, we choose $\sigma_2\!=\!0.6$, $\sigma_3\!=\!0.5$, $Z_2\!=\!5$, $Z_3\!=\!6$, and $\gamma\!=\!0.36$, as dictated by cross-validation.

\begin{figure}[t]
\centering\hspace*{-0.22cm}
\begin{subfigure}[b]{0.32\linewidth}
\centering\includegraphics[trim=0 0 0 0, clip=true, width=3.2cm]{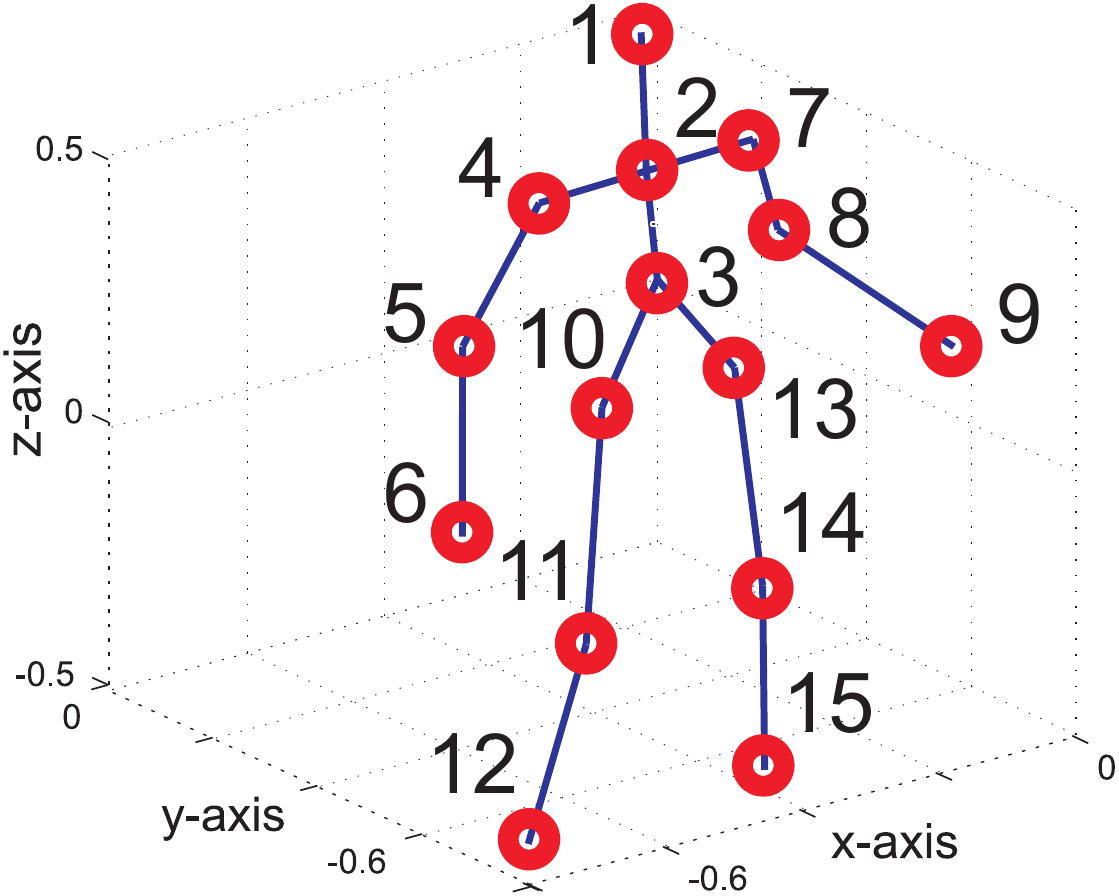}
\renewcommand{\arraystretch}{0.95}
{
\scriptsize
\begin{tabular}{ c | c | c | c | c }
\kern-0.5em A\kern-0.5em & \kern-0.5em B\kern-0.5em & \kern-0.5em C\kern-0.5em & \kern-0.5em D\kern-0.5em & \kern-0.5em E\kern-0.5em\\
\hline
\kern-0.5em 6,9\kern-0.5em & \kern-0.5em 1,6,9\kern-0.5em & \kern-0.5em 6,9,12,15\kern-0.5em & \kern-0.5em 4,6,7,9,11,14\kern-0.5em & \kern-0.5em 4,6,7,9,\kern-0.7em\\
\cline{1-4}
F & G & H & I & \kern-0.5em 11,12,\kern-0.7em\\
\kern-0.7em 4-15\kern-0.5em & \kern-0.5em 1,4-15\kern-0.5em & \kern-0.5em 1,2,4-15\kern-0.5em & \kern-0.5em 1-15\kern-0.5em & \kern-0.5em 14,15\kern-0.7em\\
\hline
\end{tabular}
}
%
%
\caption{\label{fig:piv4}}
\end{subfigure}
\hspace*{0.16cm}
\begin{subfigure}[b]{0.32\linewidth}
\centering\includegraphics[trim=0 3 0 10, clip=true, width=4.35cm]{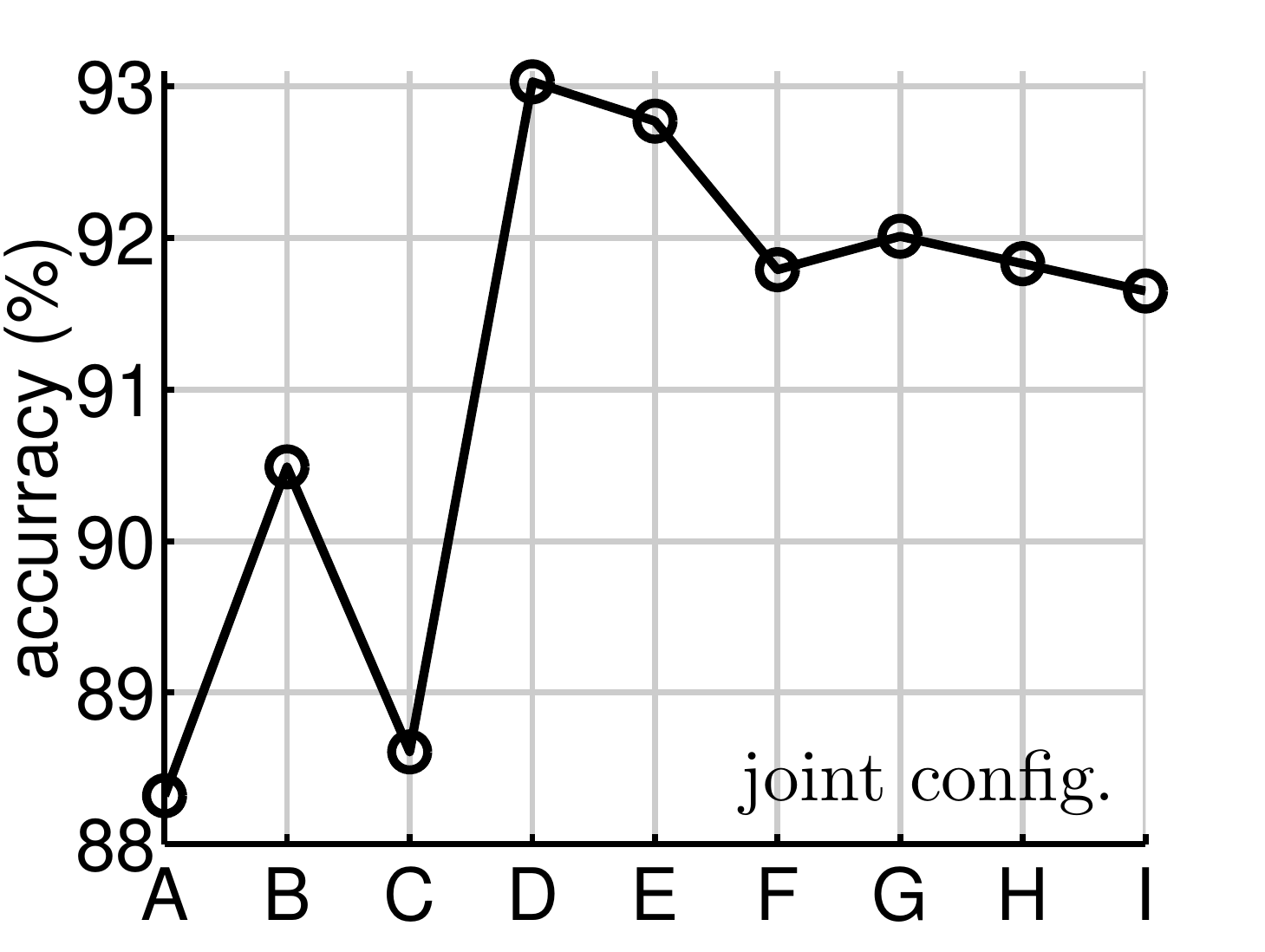}
\caption{\label{fig:piv5}}
\end{subfigure}
\hspace*{0.00cm}
\begin{subfigure}[b]{0.32\linewidth}
\centering\includegraphics[trim=0 3 0 10, clip=true, width=4.35cm]{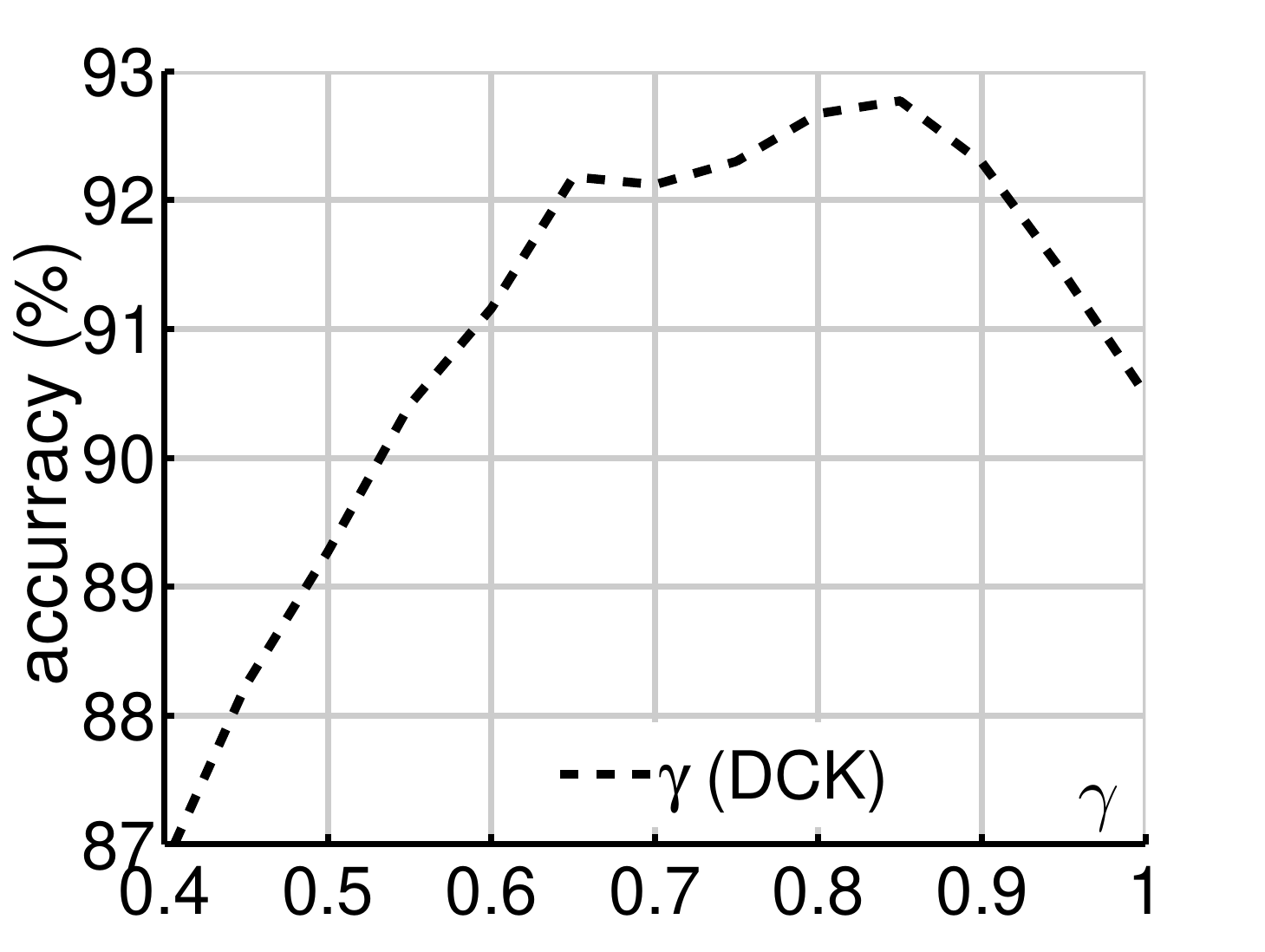}
\caption{\label{fig:piv6}}
\end{subfigure}
\caption{Figure \ref{fig:piv4} enumerates the body-joints in the Florence3D-Action dataset. The table below lists subsets A-I of the body-joints used to build  representations evaluated in Figure \ref{fig:piv5}, which demonstrates the performance of our dynamics compatibility kernel w.r.t. these subsets. Figure \ref{fig:piv6} demonstrates effectiveness of equalizing the factors in non-symmetric tensor representation by HOSVD Eigenvalue Power Normalization by varying $\gamma$.}
\end{figure}

\vspace{0.05cm}
\noindent{\textbf{Dynamics compatibility kernel.}}
In this section, we evaluate the influence of choosing parameters for the DCK kernel. Our experiments are based on the 
Florence3D-Action dataset. We present the scores on the test set as the results on the validation set match these closely. As this kernel considers all spatio-temporal co-occurrences of body-joints, we first evaluate the impact of the joint subsets we select for generating this representation as not all body-joints need to be used for describing actions.

Figure \ref{fig:piv4} enumerates the body-joints that describe every 3D human skeleton on the Florence3D-Action dataset whilst the table underneath lists the proposed body-joint subsets A-I which we use for computations of DCK. In Figure \ref{fig:piv5}, we plot the performance of our DCK kernel for each subset. The plot shows that using two body-joints associated with the hands from Configuration-A in the DCK kernel construction, we attain $88.32\%$ accuracy which highlights the informativeness of temporal dynamics. For Configuration-D, which includes six body-joints such as the knees, elbows and hands, our performance reaches $93.03\%$. This suggests that some not-selected for this configuration body-joints may be noisy and therefore detrimental to classification.

As configuration Configuration-E includes eight body-joints such as the feet, knees, elbows and hands, we choose it for our further experiments as it represents a reasonable trade-off between performance and size of representations. This configuration scores $92.77\%$ accuracy. We see that if we utilize all the body-joints according to Configuration-I, performance of $91.65\%$ accuracy is still somewhat lower compared to $93.03\%$ accuracy for Configuration-D highlighting again the issue of noisy body-joints. 

In Figure \ref{fig:piv6}, we show the performance of our DCK kernel when HOSVD factors underlying our non-symmetric tensors are equalized by varying the EPN parameter $\gamma$. For $\gamma\!=\!1$, HOSVD EPN is absent which leads to $90.49\%$ accuracy only. For the optimal value of $\gamma\!=\!0.85$, the accuracy rises to $92.77\%$. This again demonstrates the presence of the burstiness effect in temporal representations.

\vspace{0.05cm}
\noindent{\textbf{Comparison to the state of the art.}} In this section, we compare the performance of our representations against the best performing methods on the three datasets. Along with comparing SCK and DCK, we will also explore the complementarity of these representations in capturing the action dynamics by combining them.

On the Florence3D-Action dataset, we present our best results in Table \ref{tab:flor}. Note that the model parameters for the evaluation was selected by cross-validation. Linearizing a sequence compatibility kernel using these parameters resulted in a tensor representation of size $26,565$ dimensions\footnote{\label{foot:foot1}Note that this is the length of a vector per sequence after unfolding our tensor representation and removing duplicate coefficients from the symmetries in the tensor.}, and producing an accuracy of $92.98\%$ accuracy. As for the dynamics compatibility kernel (DCK), our model selected Configuration-E (described in Figure~\ref{fig:piv4}) resulting in a representation of dimensionality $16,920$ and achieved a performance of $92\%$. However, somewhat better results were attained by Configuration-D, namely $92.27\%$ accuracy for size of $9,450$.  Combining both SCK representation with DCK in Configuration-E results in an accuracy of $95.23\%$. This constitutes a $4.5\%$ improvement over the state of the art on this dataset as listed in Table \ref{tab:flor} and demonstrates the complementary nature of SCK and DCK. To the best of our knowledge, this is the highest performance attained on this dataset.

\begin{table}[t]
\renewcommand{\arraystretch}{0.85}
{
\footnotesize
\begin{subtable}[t]{0.48\linewidth}
\centering
\begin{tabular}{ c | c | c | c | c }
 & SCK & \multicolumn{2}{|c|}{DCK} & \kern-0.3em SCK+DCK\kern-0.3em\\
\hline
\kern-0.3em accuracy\kern-0.3em & \kern-0.3em $92.98\%$\kern-0.3em & \kern-0.3em $93.03\%$\kern-0.3em & \kern-0.3em $92.77\%$\kern-0.3em & \kern-0.3em $\mathbf{95.23\%}$\kern-0.3em\\
size & 26,565 & 9,450 & 16,920 & 43,485\\
\hline
\end{tabular}\vspace{0.1cm}\\\begin{tabular}{ c | c}
\hline
\kern-0.3em Bag-of-Poses $82.00\%$ \cite{seidenari_florence3d}\kern-0.3em & \kern-0.3em $SE(3)$ $90.88\%$ \cite{vemulapalli_SE3}\kern-0.3em\\
\hline
\end{tabular}
\caption{\label{tab:flor}}
\end{subtable}
\hspace{0.1cm}
\begin{subtable}[t]{0.48\linewidth}
\centering
\begin{tabular}{ c | c | c | c }
 & SCK & DCK & \kern-0.3em SCK+DCK\kern-0.3em\\
\hline
\kern-0.3em accuracy\kern-0.3em & $96.08\%$ & $97.5\%$ & $\mathbf{98.2\%}$\\
size & 40,480 & 16,920 & 57,400\\
\hline
\end{tabular}\vspace{0.1cm}\\\begin{tabular}{ c | c }
\hline
\kern-0.3em 3D joints. hist. $90.92\%$ \cite{xia_utkinect}\kern-0.3em & \kern-0.3em $SE(3)$ $97.08\%$ \cite{vemulapalli_SE3}\kern-0.3em\\
\hline
\end{tabular}
\caption{\label{tab:utk}}
\end{subtable}
}
\caption{Evaluations of SCK and DCK and comparisons to the state-of-the-art results on \ref{tab:flor} the Florence3D-Action and \ref{tab:utk} UTKinect-Action dataset.}
\end{table}

Action recognition results on the UTKinect-Action dataset are presented in Table \ref{tab:utk}. For our experiments on this dataset, we kept all the parameters the same as those we used on the Florence3D dataset (described above). On this dataset, both SCK and DCK representations yield  $96.08\%$ and $97.5\%$ accuracy, respectively. Combining SCK and DCK yields $98.2\%$ accuracy outperforming marginally a more complex approach described in~\cite{vemulapalli_SE3} which uses Lie group algebra on $SE(3)$ matrix descriptors and requires practical extensions such as discrete time warping and Fourier temporal pyramids for attaining this performance, which we avoid completely.

\begin{table}[b]
\renewcommand{\arraystretch}{0.95}
{
\footnotesize
\parbox{.48\linewidth}{
\centering	
\begin{tabular}{ c | c | c | c }
 & SCK & DCK & \kern-0.3em SCK+DCK\kern-0.3em\\
\hline
\kern-0.3em acc., prot.~\cite{wu_actionlets}\kern-0.3em & \kern-0.3em $90.72\%$\kern-0.3em & \kern-0.3em $86.30\%$\kern-0.3em & \kern-0.3em $\mathbf{91.45\%}$\kern-0.3em\\
\kern-0.3em acc., prot.~\cite{li_msraction3d}\kern-0.3em & \kern-0.3em $93.52\%$\kern-0.3em & \kern-0.3em $91.71\%$\kern-0.3em & \kern-0.3em $\mathbf{93.96\%}$\kern-0.3em\\
size & 40,480 & 16,920 & 57,400\\
\hline
\end{tabular}
}
\parbox{.48\linewidth}{
\centering
%
%
\begin{tabular}{ c | c }
\kern-0.3em accuracy, protocol~\cite{wu_actionlets}\kern-0.3em & \kern-0.3em accuracy, protocol~\cite{li_msraction3d}\kern-0.3em\\
\hline
\kern-0.3em Actionlets $88.20\%$ \cite{wu_actionlets}\kern-0.3em & \kern-0.3em R. Forests $90.90\%$ \cite{zhu_fusingjoints}\kern-0.3em\\
$SE(3)$ $89.48\%$ \cite{vemulapalli_SE3} & $SE(3)$ $92.46\%$ \cite{vemulapalli_SE3}\\
\kern-0.3em Kin. desc. $91.07\%$ \cite{zanfir_movingpose}\kern-0.3em &\\
\hline
\end{tabular}
}
}
\caption{Results on SCK and DCK and comparisons to the state of the art on MSR-Action3D.}\label{tab:msr}
\end{table}

In Table~\ref{tab:msr}, we present our results on the MSR-Action3D dataset. Again, we kept all the model parameters the same as those used on the Florence3D dataset. Conforming to prior literature, we use two evaluation protocols on this dataset, namely (i) the protocol described in actionlets~\cite{wu_actionlets}, for which the authors utilize the entire dataset with its 20 classes during the training and evaluation, and (ii) approach of~\cite{li_msraction3d}, for which the authors divide the data into three subsets and report the average in classification accuracy over these subsets. The SCK representation results in the state-of-the-art accuracy of $90.72\%$ and $93.52\%$ for the two evaluation protocols, respectively. Combining SCK with DCK outperforms other approaches listed in the table and yields $91.45\%$ and $93.96\%$ accuracy for the two protocols, respectively. 
\\
\\
\noindent{\textbf{Processing Time.}}
For SCK and DCK, processing a single sequence with unoptimized MATLAB code on a single core i5 takes 0.2s and 1.2s, respectively. Training on full MSR Action3D with the SCK and DCK takes about 13 min. In comparison, extracting $SE(3)$ features \cite{vemulapalli_SE3} takes 5.3s per sequence, processing on the full MSR Action3D dataset takes $\sim$ 50 min. and with post-processing (time warping, Fourier pyramids, etc.) it goes to about 72 min. Therefore, SCK and DCK is about $5.4\!\times$ faster. 

\section{Conclusions}\label{ref:conc}

We have presented two kernel-based tensor representations for action recognition from 3D skeletons, namely the sequence compatibility kernel (SCK) and dynamics compatibility kernel (DCK). SCK captures the higher-order correlations between 3D coordinates of the body-joints and their temporal variations, and factors out the need for expensive operations such as Fourier temporal pyramid matching or dynamic time warping, commonly used for generating sequence-level action representations. Further, our DCK kernel captures the action dynamics by modeling the spatio-temporal co-occurrences of the body-joints. This tensor representation also factors out the temporal variable, whose length depends on each sequence. Our experiments substantiate the effectiveness of our representations, demonstrating state-of-the-art performance on three challenging action recognition datasets.

\begin{appendices}
\renewcommand\title[1]{}
\newcommand\titlerunning[1]{}
\newcommand\authorrunning[1]{}
\renewcommand\author[1]{}
\newcommand\institute[1]{}
\newcommand\maketitle{}
\title{Tensor Representations via Kernel Linearization for Action Recognition from 3D Skeletons (Supp. Material)}







\maketitle

\section{Linearization of Dynamics Compatibility Kernel}
In what follows, we derive the linearization of DCK. Let us remind that $G_{\sigma}(\vu-\vub)=\exp(-\enorm{\vu - \vub}^2/{2\sigma^2})$, $G'_{\sigma}(\valpha,\vbeta)=G_{\sigma}(\valpha)G_{\sigma}(\vbeta)$ and $G_{\sigma}(\vi-\vj)=\delta(\vi-\vj)$ if $\sigma\!\rightarrow\!0$, therefore $\delta(\boldsymbol{0})=1$ and $\delta(\vu)=0$ if $\vu\!\neq\!\boldsymbol{0}$. Moreover, $\Lambda=J^2$ is a normalization constant and $\idxJ=\idx{J}\times\idx{N}$. We remind that kernel $G_{\sigma'_2}(\vx-\vy)\approx \phi(\vx)^T\phi(\vy)$ while $G_{\sigma'_3}(\frac{s-t}{N})\approx\vz(s/N)^T\vz(t/N)$. Therefore, we obtain:

{

\begin{align}
& \!K_D(\piA,\piB)=\nonumber\\
&\!\!=\!\frac{1}{\Lambda}\!\!\!\!\!\sum\limits_{\substack{(i,s)\in\idxJ\!,\\(i',s')\in\idxJ\!,\\i'\!\!\neq\!i\!,s'\!\!\neq\!s}}\sum\limits_{\substack{(\!j,t)\in\idxJ\!,\\(\!j'\!\!,t'\!)\in\idxJ,\\j'\!\!\neq\!j\!,t'\!\!\neq\!t}}\!\!\!\!G'_{\sigma'_1}(i\!-\!j\!, i'\!\!-\!j'\!)\,G_{\sigma'_2}\left(\left(\vx_{is}\!-\!\vx_{i's'}\!\right)\!-\!\left(\vy_{jt}-\vy_{j't'}\right)\right)G'_{\sigma'_3}(\frac{s-t}{N},\!\frac{s'-t'}{N})\,\cdot\nonumber\\[-24pt]
&\qquad\qquad\qquad\qquad\qquad\qquad\qquad\qquad\qquad\qquad\qquad\qquad\qquad\;\;\;\cdot G'_{\sigma'_4}(s\!-\!s'\!,t\!-\!t'\!)\nonumber\\[6pt]
&\!\!=\!\frac{1}{\Lambda}\!\!\sum\limits_{\substack{i\in\idx{J}\!,\\i'\!\in\idx{J}\!:\\i'\!\neq i}}\sum\limits_{\substack{s\in\idx{N}\!,\\s'\!\!\in\idx{N}\!:\\s'\!\!\neq\!s}}\sum\limits_{\substack{t\in\idx{N}\!,\\t'\!\!\in\idx{N}\!:\\t'\!\!\neq\!t}}\!G_{\sigma'_2}\big(\!\left(\vx_{is}\!-\!\vx_{i's'}\!\right)\!-\!\left(\vy_{jt}-\vy_{j't'}\right)\!\big)\,G_{\sigma'_3}\big(\frac{s-t}{N}\big)G_{\sigma'_3}\big(\frac{s'-t'}{N}\big)\cdot{\RaiseBiggBar{-8pt}{_{\substack{\\[-10pt]j\!=\!i\\j'\!\!=\!i'\!}}}}\nonumber\\[-24pt]
&\qquad\qquad\qquad\qquad\qquad\qquad\qquad\qquad\qquad\qquad\quad\;\cdot G_{\sigma'_4}(s\!-\!s'\!)\,G_{\sigma'_4}(t\!-\!t'\!)\nonumber\\[6pt]
&\!\!\approx\!\frac{1}{\Lambda}\!\!\sum\limits_{\substack{i\in\idx{J}\!,\\i'\!\in\idx{J}\!:\\i'\!\neq i}}\sum\limits_{\substack{s\in\idx{N}\!,\\s'\!\!\in\idx{N}\!:\\s'\!\!\neq\!s}}\sum\limits_{\substack{t\in\idx{N}\!,\\t'\!\!\in\idx{N}\!:\\t'\!\!\neq\!t}}\!\vphi\left(\vx_{is}\!-\!\vx_{i's'}\!\right)^T\!\vphi\left(\vy_{it}-\vy_{i't'}\right)\!\cdot\!\vz\big(\frac{s}{N}\big)^T\!\vz\big(\frac{t}{N}\big)\!\cdot\!\vz\big(\frac{s'\!}{N}\big)^T\!\vz\big(\frac{t'\!}{N}\big)\cdot\nonumber\\[-24pt]
&\qquad\qquad\qquad\qquad\qquad\qquad\qquad\qquad\qquad\qquad\quad\;\cdot G_{\sigma'_4}(s\!-\!s'\!)\,G_{\sigma'_4}(t\!-\!t'\!)\nonumber\\[6pt]
&\!\!=\!\frac{1}{\Lambda}\!\!\sum\limits_{\substack{i\in\idx{J}\!,\\i'\!\in\idx{J}\!:\\i'\!\neq i}}\sum\limits_{\substack{s\in\idx{N}\!,\\s'\!\!\in\idx{N}\!:\\s'\!\!\neq\!s}}\sum\limits_{\substack{t\in\idx{N}\!,\\t'\!\!\in\idx{N}\!:\\t'\!\!\neq\!t}}\!\!\left<\!G_{\sigma'_4}(s\!-\!s'\!)\left(\vphi(\vx_{is}\!\!-\!\!\vx_{i's'})
\!\cdot\!\vz\big(\frac{s}{N}\big)^T\!\right)\!\kronstack\vz\big(\frac{s'\!}{N}\big)\!\right.,\nonumber\\[-24pt]
&\qquad\qquad\qquad\qquad\qquad\qquad\qquad\qquad
\left.
G_{\sigma'_4}(t\!-\!t'\!)\Big(\vphi(\vy_{it}\!\!-\!\!\vy_{i't'})
\!\cdot\!\vz\big(\frac{t}{N}\big)^T\!\Big)\!\kronstack\vz\big(\frac{t'\!}{N}\big)\!\right>\nonumber\\
&\!\!=\!\!\!\!\sum\limits_{\substack{i\in\idx{J}\!,\\i'\!\in\idx{J}\!:\\i'\!\neq i}}
\!\!
\left<
\!\frac{1}{\sqrt{\Lambda}}\!\!\sum\limits_{\substack{s\in\idx{N}\!,\\s'\!\!\in\idx{N}\!:\\s'\!\!\neq\!s}}\!\!
G_{\sigma'_4}(s\!-\!s'\!)\left(\vphi(\vx_{is}\!\!-\!\!\vx_{i's'})
\!\cdot\!\vz\big(\frac{s}{N}\big)^T\!\right)\!\kronstack\vz\big(\frac{s'\!}{N}\big)\!\right.,\nonumber\\[-24pt]
&\qquad\qquad\qquad\qquad\qquad\!
\left.
\!\frac{1}{\sqrt{\Lambda}}\!\!\sum\limits_{\substack{t\in\idx{N}\!,\\t'\!\!\in\idx{N}\!:\\t'\!\!\neq\!t}}\!\!
G_{\sigma'_4}(t\!-\!t'\!)\,\Big(\vphi(\vy_{it}\!\!-\!\!\vy_{i't'})
\!\cdot\!\vz\big(\frac{t}{N}\big)^T\!\Big)\!\kronstack\vz\big(\frac{t'\!}{N}\big)\!\right>\!.
\label{eq:supp1}
\end{align}
Equation \eqref{eq:supp1} expresses $K_D(\piA,\piB)$ as a sum over dot-products on third-order non-symmetric tensors. We  introduce operator $\tG$ into Equation \eqref{eq:supp1} to amend the dot-product with a distance which can handle burstiness and we obtain a modified kernel:
\begin{align}
& \!\!\!\!\!\!K_D^{*}(\piA,\piB)\!
=\!\!\!\!\sum\limits_{\substack{i\in\idx{J}\!,\\i'\!\in\idx{J}\!:\\i'\!\neq i}}
\!\!
\left<
\!\tG\bigg(\!\frac{1}{\sqrt{\Lambda}}\!\!\sum\limits_{\substack{s\in\idx{N}\!,\\s'\!\!\in\idx{N}\!:\\s'\!\!\neq\!s}}\!\!
G_{\sigma'_4}(s\!-\!s'\!)\,\left(\vphi(\vx_{is}\!\!-\!\!\vx_{i's'})
\!\cdot\!\vz\big(\frac{s}{N}\big)^T\!\right)\!\kronstack\vz\big(\frac{s'\!}{N}\big)\!\right.\!\!\bigg),\nonumber\\[-6pt]
&\qquad\qquad\qquad\qquad\!\!
\tG\bigg(\!\left.
\!\frac{1}{\sqrt{\Lambda}}\!\!\sum\limits_{\substack{t\in\idx{N}\!,\\t'\!\!\in\idx{N}\!:\\t'\!\!\neq\!t}}\!\!
G_{\sigma'_4}(t\!-\!t'\!)\,\Big(\vphi(\vy_{it}\!\!-\!\!\vy_{i't'})
\!\cdot\!\vz\big(\frac{t}{N}\big)^T\!\Big)\!\kronstack\vz\big(\frac{t'\!}{N}\big)\!\bigg)\!\right>\!.\!\!
\label{eq:supp2}
\end{align}
From Equation \eqref{eq:supp2} the following notation is introduced:
\begin{align}
& \!\!\!\!\tV_{ii'\!}\!=\!\mygthree{\tX_{ii'\!}}\!,\!\text{ and }\tX_{ii'\!}\!=\!\!\frac{1}{\sqrt{\Lambda}}\!\!\sum\limits_{\substack{s\in\idx{N}\!,\\s'\!\!\in\idx{N}\!:\\s'\!\!\neq\!s}}\!\!
G_{\sigma'_4}(s\!-\!s'\!)\left(
\vphi(\vx_{is}\!\!-\!\!\vx_{i's'})
\!\cdot\!\vz\big(\frac{s}{N}\big)^T\!\right)\!\kronstack\vz\big(\frac{s'\!}{N}\big),
\label{eq:supp3}
\end{align}
where the summation over the pairs of body-joints in Equation \eqref{eq:supp2} is replaced by the concatenation along the fourth mode to obtain representations $\big[\tV_{ii'\!}\big]_{i>i'\!:\,i,i'\in\idx{J}}^{\oplus_4}$ and $\big[\tVH_{ii'\!}\big]_{i>i'\!:\,i,i'\in\idx{J}}^{\oplus_4}$ for $\piA$ and $\piB$. Thus, $K_D^{*}$ becomes:
\begin{align}
& K_D^{*}(\piA,\piB)=\left<\sqrt{2}\big[\tV_{ii'\!}\big]_{i>i'\!:\,i,i'\in\idx{J}}^{\oplus_4}\!,\sqrt{2}\big[\tVH_{ii'\!}\big]_{i>i'\!:\,i,i'\in\idx{J}}^{\oplus_4}\right>\label{eq:supp4}
\end{align}
As Equation \eqref{eq:supp4} suggests, we avoid repeating the same computation when evaluating our representations \eg, we stack only unique pairs of body-joints $i\!>\!i'\!$. Moreover, we also ensure we execute computations temporally only for $s\!>\!s'\!$. In practice, we have to evaluate only $\binom{JN}{2}$ unique spatio-temporal pairs in Equation \eqref{eq:supp4} rather than naive  $J^2N^2$ per sequence. The final representation is of $Z'_2\!\cdot\!\binom{JZ'_3\!}{2}$ size, where $Z'_2$ and $Z'_3$ are the numbers of pivots for approximation of $G_{\sigma'_2}\!$ and $G_{\sigma'_3}$. 

We assume that all sequences have $N$ frames for simplification of presentation. Our formulations are equally applicable to sequences of arbitrary lengths \eg,~$M$ and $N$. Thus, we apply in practice $G'_{\sigma'_3}(\frac{s}{M}\!-\!\frac{t}{N},\frac{s'}{M}\!-\!\frac{t'}{N})$ and $\Lambda\!=\!J^2MN$ in Equation \eqref{eq:supp1}.

In practice, we use $G^{'}_{\sigma'_2}(\vx\!-\!\vy)\!=\!G_{\sigma'_2}(x^{(x)}\!\!-\!y^{(x)})\!+\!G_{\sigma'_2}(x^{(y)}\!\!-\!y^{(y)})\!+\!G_{\sigma'_2}(x^{(z)}\!\!-\!y^{(z)})$ so the kernel $G^{'}_{\sigma'_2}(\vx\!-\!\vy)\approx[\phi(x^{(x)}\!); \phi(x^{(y)}\!); \phi(x^{(z)}\!)]^T\![\phi(y^{(x)}\!); \phi(y^{(y)}\!); \phi(y^{(z)}\!)]$ but for simplicity we write $G_{\sigma'_2}(\vx\!-\!\vy)\!\approx\!\phi(\vx)^T\phi(\vy)$. Note that $(x)$, $(y)$, $(z)$ are the spatial xyz-components of displacement vectors \eg, $\vx_{is}\!-\!\vx_{i's'}$.

\section{Positive Definiteness of SCK and DCK}
SCK and DCK utilize sums over products of RBF subkernels. It is known from \cite{taylor_kermet} that sums, products and linear combinations (for non-negative weights) of positive definite kernels result in positive definite kernels.

Moreover, subkernel $G_{\sigma'_2}\left(\left(\vx_{is}\!-\!\vx_{i's'}\!\right)\!-\!\left(\vy_{jt}-\vy_{j't'}\right)\right)$ employed by DCK in Equation \eqref{eq:supp1} (top) can be rewritten as:
\begin{align}
& G_{\sigma'_2}\left(\vz_{isi's'}\!-\!\vz'_{jtj't'}\!\right)\text{ where }\vz_{isi's'}\!=\!\vx_{is}\!-\!\vx_{i's'}\text{ and }\vz'_{jtj't'}\!=\!\vy_{jt}-\vy_{j't'}.\label{eq:supp5}
\end{align}

The RBF kernel $G_{\sigma'_2}$ is positive definite by definition and the mappings from $\vx_{is}$ and $\vx_{i's'}$ to $\vz_{isi's'}$ and from $\vy_{jt}$ and $\vy_{j't'}$ to $\vz'_{jtj't'}$, respectively, are unique. Therefore, the entire kernel is positive definite.

Lastly, whitening performed on SCK also results in a positive (semi)definite kernel as we employ the Power-Euclidean kernel \eg, if $\mX$ is PD then $\mX^\gamma$ stays also PD for $0\!<\!\gamma\!\leq\!1$ because $\mX^\gamma\!=\!\mU\mLambda^\gamma\mV$ and element-wise rising of eigenvalues to the power of $\gamma$ gives us $\diag(\mLambda)^\gamma\!\geq\!0$. Therefore, the sum over dot-products of positive (semi)definite autocorrelation matrices raised to the power of $\gamma$ remains positive (semi)definite.

\section{Complexity} 
Non-linearized SCK with kernel SVM have complexity $\bigoh(JN^2T^\rho)$ given $J$ body joints, $N$ frames per sequence, $T$ sequences, and $2\!<\!\rho\!<\!3$ which concerns complexity of kernel SVM. Linearized SCK with linear SVM have complexity $\bigoh(JNTZ_*^r)$ for total of $Z_*$ pivots and tensor of order $r\!=\!3$. As $N^2T^\rho\!\gg\!NTZ_*^r$. For $N\!=\!50$ and $Z_*\!=\!20$, \eg~$Z_*\!=\!3Z_2\!+\!Z_3$ given $G_{\sigma'_2}$ and $G_{\sigma'_3}$, linearization is $3.5\!\times$ (or $32\!\times$) faster than the exact kernel if $T\!=\!557$ (or $T\!=\!5000$, respectively).

Non-linearized DCK with kernel SVM have complexity $\bigoh(J^2N^4T^\rho)$. Linearized DCK with SVM enjoys $\bigoh(J^2N^2TZ^3)$ for $Z$ pivots per kernel, \eg~$Z\!=\!Z_2\!=\!Z_3$ given $G_{\sigma'_2}$ and $G_{\sigma'_3}$. As $N^4T^\rho\!\gg\!N^2TZ^3$, the linearization is $~11000\!\times$ faster than the exact kernel, for say $Z\!=\!5$. 

Slice-wise EPN applied to SCK has negligible cost  $\bigoh(JTZ_*^{\omega+1})$, where $2\!<\!\omega\!<\!2.376$ concerns complexity of eigenvalue decomposition applied to each tensor slice.

EPN applied to DCK utilizes HOSVD and results in complexity $\bigoh(J^2TZ^4)$. As HOSVD is performed by truncated SVD on matrices obtained from unfolding $\tV_{ii'\!}\in\mbr{Z\times Z\times Z}\!$ along a chosen mode, $\bigoh(Z^4)$ represents the complexity of truncated SVD on matrices $\mV_{ii'\!}\in\mbr{Z\times Z^2}\!$ which can attain rank less or equal to $Z$.

\comment{
\section{Third-order Statistics -- illustration} 

\begin{figure}[b]
\centering
\includegraphics[trim=0 3 0 25, clip=true, width=4.7cm]{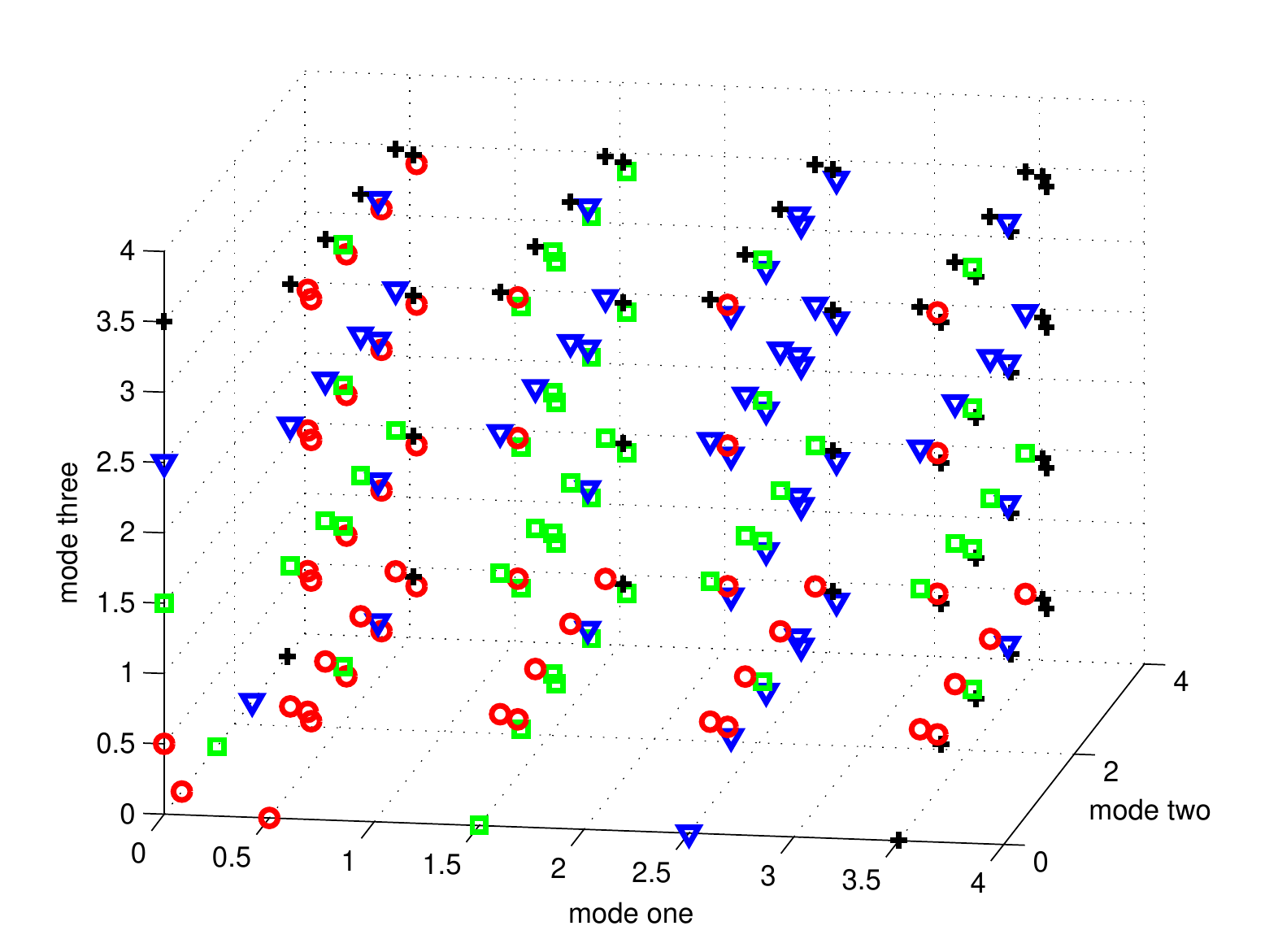}
%
\caption{Illustration of third-order statistics in SCK. Linearized components $\vphi(x)$, $\vphi(y)$, $\vphi(z)$ and time $\vz(t)$ denoted as ($\circ$, $\scriptscriptstyle\square$, $\scriptstyle\triangledown$, $\scriptscriptstyle+$) are captured in third-order tensor \eg, triplets ($\circ\scriptscriptstyle\square\scriptstyle\triangledown$) and ($\circ\scriptscriptstyle\square\scriptscriptstyle+$). This exposes SVM to  rich body-joint statistics. 
}
\end{figure}
}
}
\end{appendices}

{\small

}

\end{document}